**Budapest University of Technology and Economics**

Faculty of Electrical Engineering and Informatics

Department of Automation and Applied Informatics

# Developing neural machine translation models for Hungarian-English

Bachelor's Thesis

*Author*

Attila Nagy

*Advisor*

Judit Ács

May 23, 2021

# Contents

# HALLGATÓI NYILATKOZAT

Alulírott *Nagy Attila*, szigorló hallgató kijelentem, hogy ezt a szakdolgozatot meg nem engedett segítség nélkül, saját magam készítettem, csak a megadott forrásokat (szakirodalom, eszközök stb.) használtam fel. Minden olyan részt, melyet szó szerint, vagy azonos értelemben, de átfogalmazva más forrásból átvettem, egyértelműen, a forrás megadásával megjelöltem.

Hozzájárulok, hogy a jelen munkám alapadatait (szerző(k), cím, angol és magyar nyelvű tartalmi kivonat, készítés éve, konzulens(ek) neve) a BME VIK nyilvánosan hozzáférhető elektronikus formában, a munka teljes szövegét pedig az egyetem belső hálózatán keresztül (vagy autentikált felhasználók számára) közzétegye. Kijelentem, hogy a benyújtott munka és annak elektronikus verziója megegyezik. Dékáni engedéllyel titkosított diplomatervek esetén a dolgozat szövege csak 3 év eltelte után válik hozzáférhetővé.

Budapest, 2021. május 23.

*Nagy Attila*
hallgató

# Kivonat

Munkámban neurális gépi fordító modelleket tanítok be a magyar-angol, illetve angol-magyar nyelvpárokra a Hunglish2 korpusz segítségével. A dolgozat fő kontribúciója különféle adat augmentációs módszerek kiértékelése a fordító modellek tanítása során. Öt különböző augmentációs technikát mutatok be, amelyek ahelyett, hogy véletlenszerűen kicserélnének vagy kitakarnának szavakat, a mondatok dependenciaelemzése segítségével augmentálják a mondatokat. Részletes áttekintést adok a neurális hálók, szekvencia modellezés, neurális gépi fordítás, dependenciaelemzés, illetve adat augmentációs technikák elméleti hátteréről. Ezután bemutatom a Hunglish2 korpuszon végzett feltáró adatelemzés, illetve előfeldolgozás eredményeit. A legjobb magyar-angol fordító modell 33.9-es, míg a legjobb angol-magyar modell 28.6-os BLEU értéket ért el.

# Abstract

I train models for the task of neural machine translation for English-Hungarian and Hungarian-English, using the Hunglish2 corpus. The main contribution of this work is evaluating different data augmentation methods during the training of NMT models. I propose 5 different augmentation methods that are structure-aware, meaning that instead of randomly selecting words for blanking or replacement, the dependency tree of sentences is used as a basis for augmentation. I start my thesis with a detailed literature review on neural networks, sequential modeling, neural machine translation, dependency parsing and data augmentation. After a detailed exploratory data analysis and preprocessing of the Hunglish2 corpus, I perform experiments with the proposed data augmentation techniques. The best model for Hungarian-English achieves a BLEU score of 33.9, while the best model for English-Hungarian achieves a BLEU score of 28.6.

# Chapter 1

# Introduction

With the advent of deep learning, an increasing amount of problems can be solved in a data-driven manner. This applies to a large number of challenges in natural language processing as well, including machine translation (MT). The demand for creating systems that are capable of translating sentences between arbitrary languages has been around longer than the deep learning revolution. There are thousands of languages spoken around the world and therefore the number of practical applications for machine translation is very high. From automatically translating websites to translating legal documents to hundreds of languages in the European Union, automatic translation has become parts of our lives. Even though there already exist models that are capable of producing human-level translations, MT still has many unsolved challenges, e.g. dealing with long sentences and rare words, domain-specific translation models, the amount of data needed for training, just to mention a few. For these reasons, machine translation is a very actively researched area in NLP today.

Early machine translation systems relied on hand-crafted rules based on our knowledge of linguistics, but unfortunately these could not cover all the exceptions in the different languages. As time went by and an increasing amount of data was available, statistical machine translation (SMT) came into prominence. Although this is a data-driven approach, it struggles to deal with long-distance dependencies, which is essential for creating high-quality translation models. The remedy for this came with neural networks and deep learning, particularly with the transformer architecture. With neural machine translation (NMT), it is possible to capture complex and long-range dependencies, but to do so generally requires very large amounts of data (usually millions of data points). Because of this, approaches that lessen the need for collecting large, high-quality parallel corpora are valued greatly (e.g. unsupervised machine translation or data augmentation).

The contribution of my thesis is twofold. Firstly, there has been little focus on publishing neural machine translation models for Hungarian-English and English-Hungarian. That being the case, one main objective of this work, is to develop models based on the most recent architectures, that are capable of generating close to human-level translations for the two language pairs mentioned above. Secondly, data augmentation in NMT in previous works usually disregard syntactic relations in a sentence. With that in mind, I propose 5 structure-aware data augmentation techniques and evaluate them alongside the baseline model for both language pairs.

The structure of my thesis is as follows: in Chapter 2, I give a detailed overview of the fundamental mathematics of neural networks. I put particular emphasis on sequence-to-sequence learning as it is essential for machine translation. I describe the inner workings of the transformer model that are used during all my experiments. I summarize the development of NMT architectures and define the task as a sequence-to-sequence problem. I also give a brief introduction to dependency parsing as this serves as a basis for the proposed methods. I conclude this chapter with highlighting the importance of data augmentation for machine learning and by reviewing recent works in data augmentation for NLP. In Chapter 3, I describe the proposed structure-aware data augmentation methods in detail. I also discuss the implementation details of the experiments, including exploratory data analysis, preprocessing, data augmentation and model training. In Chapter 4, I present the results of the experiments and discuss my findings. Based on these, I propose some ideas for future work. In Chapter 5, I summarize and conclude my thesis.

# Chapter 2

# Background

## 2.1 Neural networks

In the following section, I will give an introduction to the mathematics of neural networks primarily based on the Deep Learning textbook by Ian Goodfellow, Yoshua Bengio and Aaron Courville [15].

### 2.1.1 Feedforward neural networks

A feedforward network is a computational model inspired by biological neural networks. The purpose of such a model is to approximate a function $f$ for a given input vector $\mathbf{x}$ as:

$$\hat{y} = f(\mathbf{x}; \boldsymbol{\theta}), \tag{2.1}$$

where $\boldsymbol{\theta}$ denotes the learnable parameters of the network. The model is a composition of $n$ functions (layers):

$$f(\boldsymbol{x}) = f^{(n)}(f^{(n-1)}...(f^{(2)}(f^{(1)}(\mathbf{x})))), \tag{2.2}$$

where the output of the $n-1$th layer serves as the input of the $n$th layer. By definition, feedforward networks restrict feedback connections, such as directed cycles or loops in the computational graph. Consequently the model can be represented as a directed acyclic graph (as shown in figure 2.1).

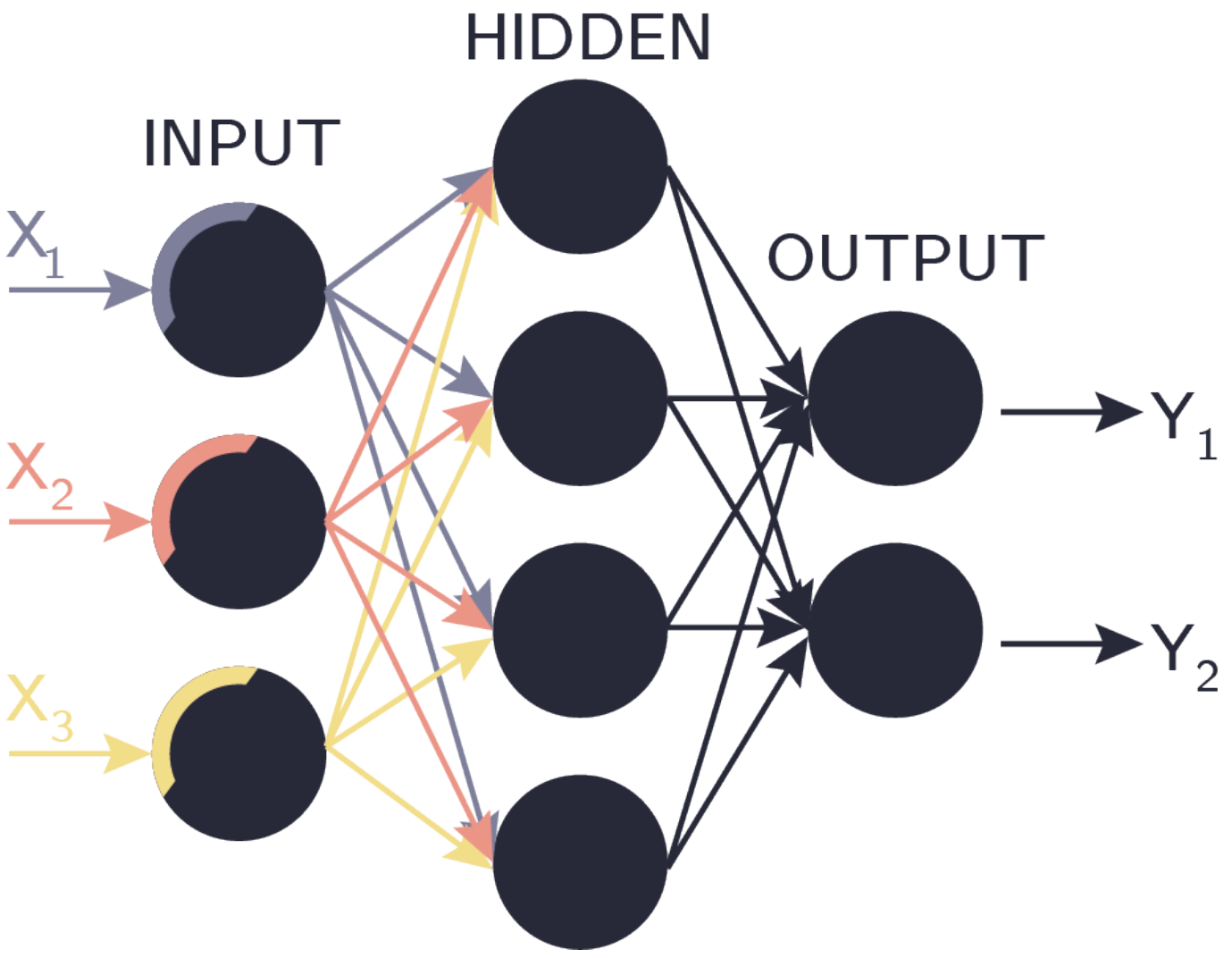

**Figure 2.1:** A neural network as a directed acyclical graph

The simplest architecture where $n = 1$ is called a single-layer perceptron. In this case we are given an input vector $\mathbf{x}$ for which we compute:

$$\hat{y} = \boldsymbol{\theta}\mathbf{x} + b, \tag{2.3}$$

where $\boldsymbol{\theta}$ are the learnable weights in the singular layer and $b$ denotes the bias. This by far is a linear model, which can only approximate linear functions, therefore in order to extend the set of representable functions, a $\phi$ nonlinearity is added to the model in the form of an activation function:

$$z = \phi(\boldsymbol{\theta}\mathbf{x} + b) \tag{2.4}$$

The Universal Approximation Theorem [20, 9] states that a feedforward network having a linear output layer and at least one hidden layer squashed through an activation function is capable of approximating even nonlinear functions. This however is challenging in practice as the single layer may have an infeasible size and therefore the algorithm may fail to generalise. In practice, connecting multiple relatively narrow layers turned out to be efficient, because it introduces levels of abstraction to the representation of knowledge [30]. Recent advances in machine learning rely heavily on deep learning: the science of designing neural networks, where the number of layers in the model is considered high.

In order to find the mapping between the input and output of the network, firstly a loss function needs to be defined between the predicted output $\hat{y}$ and the actual output $y$.

During the training of a neural network, this is what we intend to minimise by computing the gradient w.r.t. the loss function:

$$\boldsymbol{\theta_{t+1}} \leftarrow \boldsymbol{\theta_t} - \alpha \nabla_{\boldsymbol{\theta}} L(\boldsymbol{\theta}), \tag{2.5}$$

where $\alpha$ is a positive-definite step size (learning rate) and $L(\boldsymbol{\theta})$ is the defined loss function. This unfortunately applies for one training example, so in order to compute the gradients for a whole iteration through the training data, we need to calculate the gradients for every data point and take the average of the calculated gradients. The data used for training may be huge in size to induce generalization, perhaps consisting of millions of examples, therefore the proposed gradient descent in it's form is computationally very expensive to perform. In practice, an extended version of gradient descent is used, namely stochastic gradient descent (SGD). The motivation behind SGD, is that a gradient can be closely estimated using only a small subset of the training samples, so for every computation of the gradient, a minibatch of examples are sampled uniformly from the training set. This way the gradient computes as:

$$\boldsymbol{g} = \frac{1}{m} \nabla_{\boldsymbol{\theta}} \sum_{i=1}^{m} L(\boldsymbol{\theta}), \tag{2.6}$$

where $m$ denotes the size of a minibatch. The gradient then computes accordingly as:

$$\boldsymbol{\theta_{t+1}} \leftarrow \boldsymbol{\theta_t} - \alpha \boldsymbol{g}. \tag{2.7}$$

The parameter updating algorithm can be further improved by using the momentum method to accelerate convergence and avoid situations where the gradient is stuck in local minima by including previously calculated gradients into the computation at the current time step as:

$$\boldsymbol{v_{t+1}} = \gamma \boldsymbol{v_t} - \alpha \boldsymbol{g} \tag{2.8}$$

$$\boldsymbol{\theta_{t+1}} = \boldsymbol{\theta_t} + \boldsymbol{v_{t+1}}, \tag{2.9}$$

where $\boldsymbol{v}$ is the velocity parameter containing the weighted sum of all previous gradients and $\gamma$ is the decaying factor. Even more frequently used in practice is a modified version of the momentum method, called Nesterov momentum. It introduces one important addition to the previously discussed method, which is related to the calculation of gradients. When calculating $\boldsymbol{\theta}_{t+1}$, instead of taking $\boldsymbol{\theta_t}$ as a starting point in the parameter space, we compute the gradient term from an intermediate point $\boldsymbol{\theta_{intermediate}} = \boldsymbol{\theta_t} + \gamma \boldsymbol{v_t}$. The updated expression for the Nesterov accelerated gradient is:

$$\boldsymbol{v_{t+1}} = \gamma \boldsymbol{v_t} - \alpha \nabla L(\boldsymbol{\theta_t} + \gamma \boldsymbol{v_t}) \tag{2.10}$$

$$\boldsymbol{\theta_{t+1}} = \boldsymbol{\theta_t} + \boldsymbol{v_{t+1}} \tag{2.11}$$

The underlying reason for this is that in some cases, the parameter updates might result in a significant increase in the loss function and this information remains encoded in the velocity vector during computations in the future. This results in oscillations, which could significantly slow down the convergence towards the minimum.

As the predicted output is a linear combination of nonlinear terms composed of the input data, the weights and the biases, we actually compute the partial derivatives of the loss function w.r.t. each parameter with backpropagation [45]. This is done via the recursive application of the chain rule.

The selection of activation functions have a huge influence on the convergence of the learning algorithm. When relying on the Universal Approximation Theorem, it is important to note, that it only holds for bounded nonlinearities. These nonlinear functions have a compact range as they squeeze the input values into a bounded subset of the set of real numbers, hence they are often referred to as saturating functions. Traditionally, the functions sigmoid as $\phi_s(x) = \frac{1}{1+e^{-x}}$ (Figure 2.2 left) and hyperbolic tangent as $\phi_{ht}(x) = \frac{e^x - e^{-x}}{e^x + e^{-x}}$ (Figure 2.2 middle) were used in neural networks.

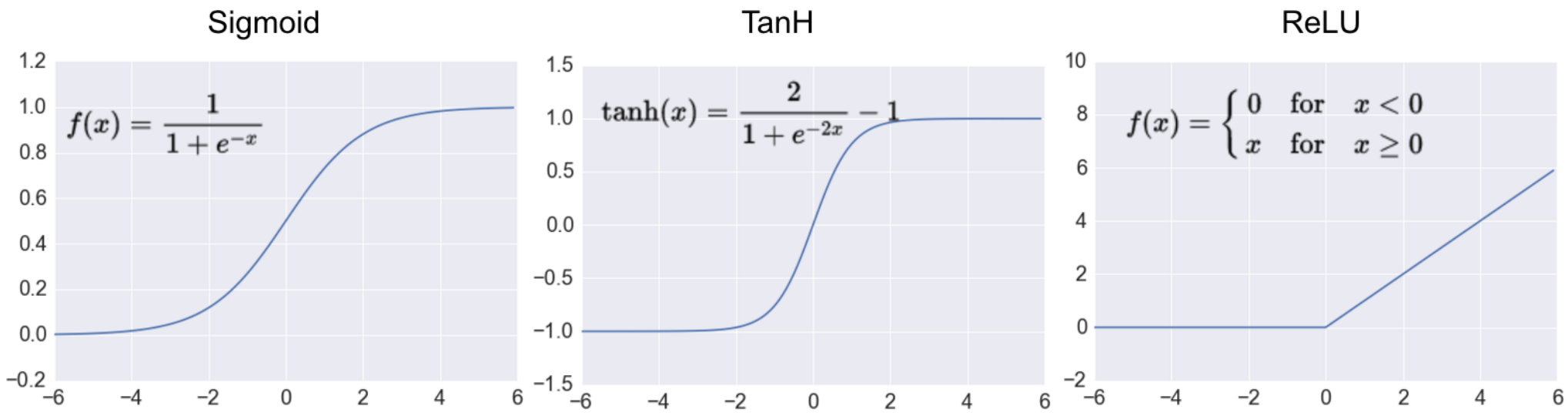

**Figure 2.2:** The three commonly used activation functions, sigmoid (left), hyperbolic tangent (middle) and the rectified linear unit (right) [35]

These however, cause some difficulties in practice. On one hand, the derivative of these functions are close to zero at a significant number of points, which slow down gradient descent. On the other hand, gradients could vanish during optimization, due to the fact that the absolute value produced by these saturating functions is smaller than one. This is especially crucial in deep neural networks, because when the gradient of the error is propagated backwards through the network, the further a parameters is from the output layer, the harder it is to train.

One significant milestone in deep learning research was the introduction of an unbound nonlinearity, namely the rectified linear unit (ReLU) as $\phi_r(x) = max(0, x)$ [37] (Figure 2.2 right). As the positive values are not downscaled during training, ReLU makes the learning algorithm more robust against vanishing gradients. It is worth noting, that ReLU is non-differentiable at $x = 0$, but conventionally it is considered as 0, which fortunately does not cause complications in practice.

### 2.1.2 Recurrent neural networks

Time is a unique dimension in the context of neural networks as it cannot be treated the same way as any other dimension of the data. Due to the fact that time is strictly progressing in one direction, classical neural networks are unable to efficiently capture the temporal representations in sequential data. However, *Recurrent Neural Networks* [45] were created with the purpose of solving this problem by enabling recurrent connections (loops) in the topology of neural networks. During every iteration, the RNN uses an input vector $\boldsymbol{x_t}$ and a hidden state vector $\boldsymbol{h_{t-1}}$ to compute $\boldsymbol{h_t}$ with the following recursive formula:

$$\boldsymbol{h}_t = f(\boldsymbol{h}_{t-1}, \boldsymbol{x}_t; \boldsymbol{\theta}), \tag{2.12}$$

where $\boldsymbol{\theta}$ denotes the learnable parameters of the network. As the output of the previous time step serve as the input of the current one, the computational graph can simply be unfolded (as shown on figure 2.3), on which we perform back-propagation through time (BPTT) [55] to compute the gradients during training.

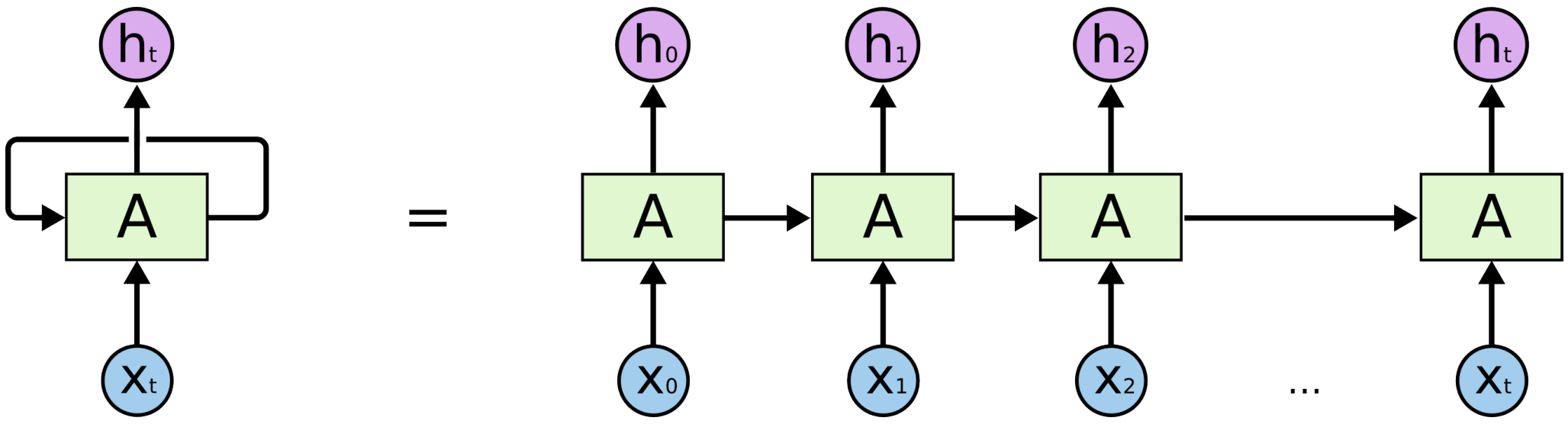

**Figure 2.3:** The unrolled computations in an RNN [39]

Theoretically, recurrent neural networks are capable of learning dependencies of any length, however it has been shown [3] that in practice the problem of vanishing and exploding gradients make it very difficult to learn long-term dependencies.

### 2.1.3 Long Short-Term Memory networks

A *Long Short-Term Memory Network* (LSTM) [19] is a special kind of RNN capable of overcoming the major shortcoming of simple recurrent neural networks, the problem of long-term dependencies by introducing self-loops. This augmentation prevents exploding/vanishing gradients by enabling them to flow for long durations. Besides the external recurrencies used by standard recurrent neural networks, the self-loops of long short-term memory networks are located internally in the LSTM cells. The inputs and outputs of an LSTM and a standard RNN are similar in structure, the crucial difference lies in the more complex internal computations of the LSTM cells. Here, the core component is called the state unit, which is governed by other internal gates, such as the forget gate, input gate and output gate. The forget gate is characterized by the following equation:

$$f_i^{(t)} = \sigma\left(b_i^f + \sum_j U_{i,j}^f x_j^{(t)} + \sum_j W_{i,j}^f h_j^{(t-1)}\right) \tag{2.13}$$

where $\sigma(\cdot)$ computes an element-wise sigmoid function, $\boldsymbol{x}^{(t)}$ and $\boldsymbol{h}^{(t)}$ denote the input vector and the hidden layer vector at time step $t$. The variables $\boldsymbol{b}^f$, $\boldsymbol{U}^f$ and $\boldsymbol{W}^f$ respectively are the biases, input weights and recurrent weights corresponding to the forget gate. The input gate is structured similarly:

$$g_i^{(t)} = \sigma\left(b_i^g + \sum_j U_{i,j}^g x_j^{(t)} + \sum_j W_{i,j}^g h_j^{(t-1)}\right) \tag{2.14}$$

having it's own weights and biases respectively denoted by $\boldsymbol{b}^g$, $\boldsymbol{U}^g$ and $\boldsymbol{W}^g$. Governed by the components introduced above, the state unit is denoted with $s_i^{(t)}$ and given by the following equation:

$$s_i^{(t)} = f_i^{(t)} s_i^{(t-1)} + g_i^{(t)} \sigma\left(b_i + \sum_j U_{i,j} x_j^{(t)} + \sum_j W_{i,j} h_j^{(t-1)}\right), \tag{2.15}$$

where $\boldsymbol{b}, \boldsymbol{U}$ and $\boldsymbol{W}$ respectively mark the biases, input weights and recurrent weights of the LSTM cell. At last, the state unit is passed through a hyperbolic tangent nonlinearity, thus the output is given as:

$$h_i^{(t)} = tanh(s_i^{(t)})q_i^{(t)}, \tag{2.16}$$

where $q_i^{(t)}$ is responsible for gating the output as:

$$q_i^{(t)} = \sigma\left(b_i + \sum_j U_{i,j}x_j^{(t)} + \sum_j W_{i,j}h_j^{(t-1)}\right), \tag{2.17}$$

with $\boldsymbol{b}^o$, $\boldsymbol{U}^o$ and $\boldsymbol{W}^o$ denoting the biases, input weights and recurrent weights.

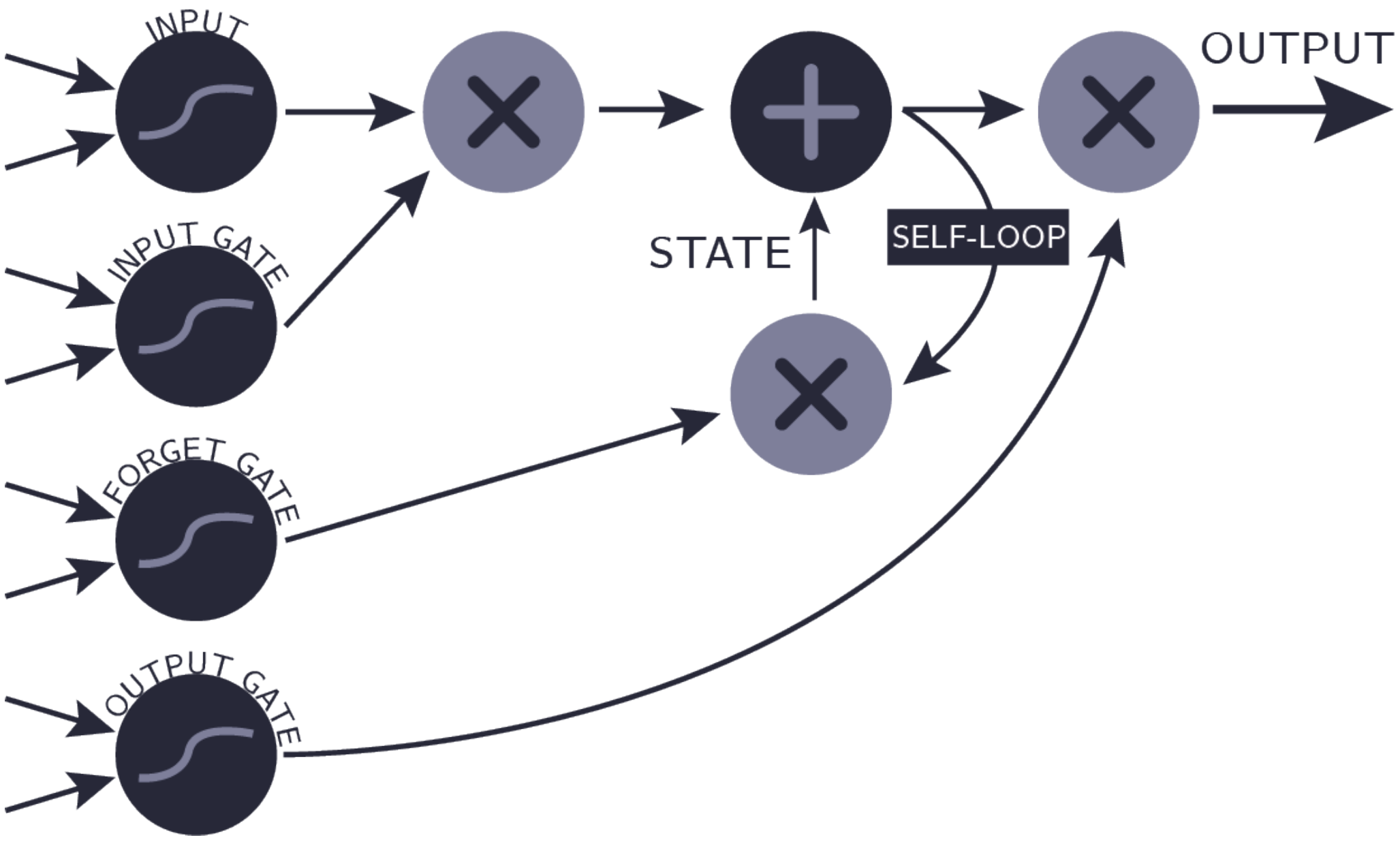

**Figure 2.4:** The internal structure of an LSTM cell

### 2.1.4 Sequence to sequence models

The previously introduced sequential models are capable of mapping input sequences to a fixed-size vector and vice versa. RNNs can also map an input sequence to a given length and output a sequence of the same length. This however, constrains their applicability in practice. Many applications, such as question answering, speech recognition or machine translation demand that the length of the input and output sequences processed by the model are different. In this section I discuss how a recurrent neural network can be trained to learn a mapping between sequences of different length. Such architectures are called sequence to sequence models or encoder-decoder networks [49, 8]. The core idea of these models is that the encoder RNN takes an input sequence and maps it to a latent vector representation $v$. This vector (often called the context vector) represents the semantic summary of the input sequence, which is then fed into the decoder, which produces the output sequence token by token. Formally, given an input sequence $x_1, ..., x_{\hat{T}}$ and output sequence $y_1, ..., y_T$ a standard sequence generation using a sequence-to-sequence architecture can be summarized in the following equation:

$$p(y_1, ..., y_T | x_1, ..., x_{\hat{T}}) = \prod_{t=1}^{\hat{T}} p(y_t | v, y_1, .., y_{t-1}) \tag{2.18}$$

Note that $T$ and $\hat{T}$ might not be equal. Similarly to recurrent neural networks, the training of sequence-to-sequence models can be performed by maximizing the following expression:

$$\frac{1}{\mathbf{S}} \sum_{T,S \in \mathbf{S}} log(p(T|S)) \tag{2.19}$$

where $\mathbf{S}$ denote the training set and $S$ and $T$ denote the source and target sequences respectively. Backpropagation through time is similarly applicable through the whole model, allowing the joint optimization of weights in the encoder and the decoder using an optimization technique, e.g. stochastic gradient descent.

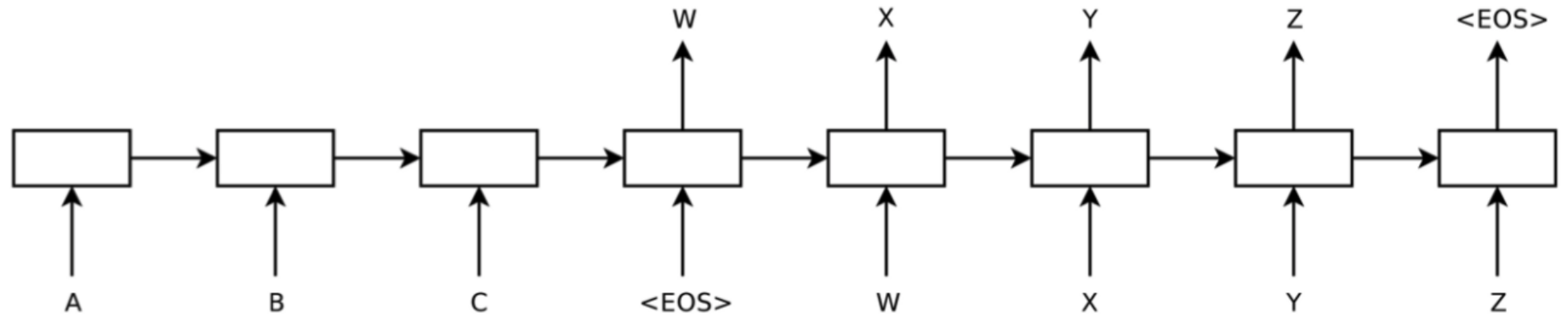

**Figure 2.5:** The sequence-to-sequence architecture [49]

Figure 2.5 illustrates the defined architecture. $A, B$ and $C$ serve as the input to the encoder, with $W, X, Y$ and $Z$ being the output tokens. The usage of the $< eos >$ (end of sentence) token is key in allowing the generation of sequences with arbitrary length. The generation of actual output sequences from the probability distribution of words can be done in multiple ways. A straight forward approach is to take the output of the softmax function and select the word with the highest probability. This method is deterministic, because it always produces the same output sequence for the same input sequence. Unfortunately however, there is no guarantee that such a greedy search yields the optimal sequence. Decoding can also be done in a stochastic manner, e.g. using Roulette Wheel Selection [34], where each word can be generated from the output of the softmax function timestep to timestep. Although this being feasible, it is better to perform the decoding at once, by taking the sequence, which corresponds to the highest probability.

$$\hat{T} = \max_{T} p(T|S) \tag{2.20}$$

where $S$ and $T$ denote the source and target sequences respectively. The computational complexity of performing an exhaustive search to get the optimal sequence is excessively high as it scales exponentially with the length of the output sequence. In practice, to make the computations tractable, decoding is usually performed via a left-to-right beam search [17]. It is characterized by a hyperparameter named beam size, denoted with $k$. At the first time step, $k$ tokens are retained, having the highest conditional probabilities. These will respectively serve as the first token of $k$ candidate output sequences. In the following time steps, $k$ tokens are iteratively selected that have the highest conditional probabilities, based on the $k$ candidate output sequences from the previous time steps. The process is illustrated in Figure 2.6.

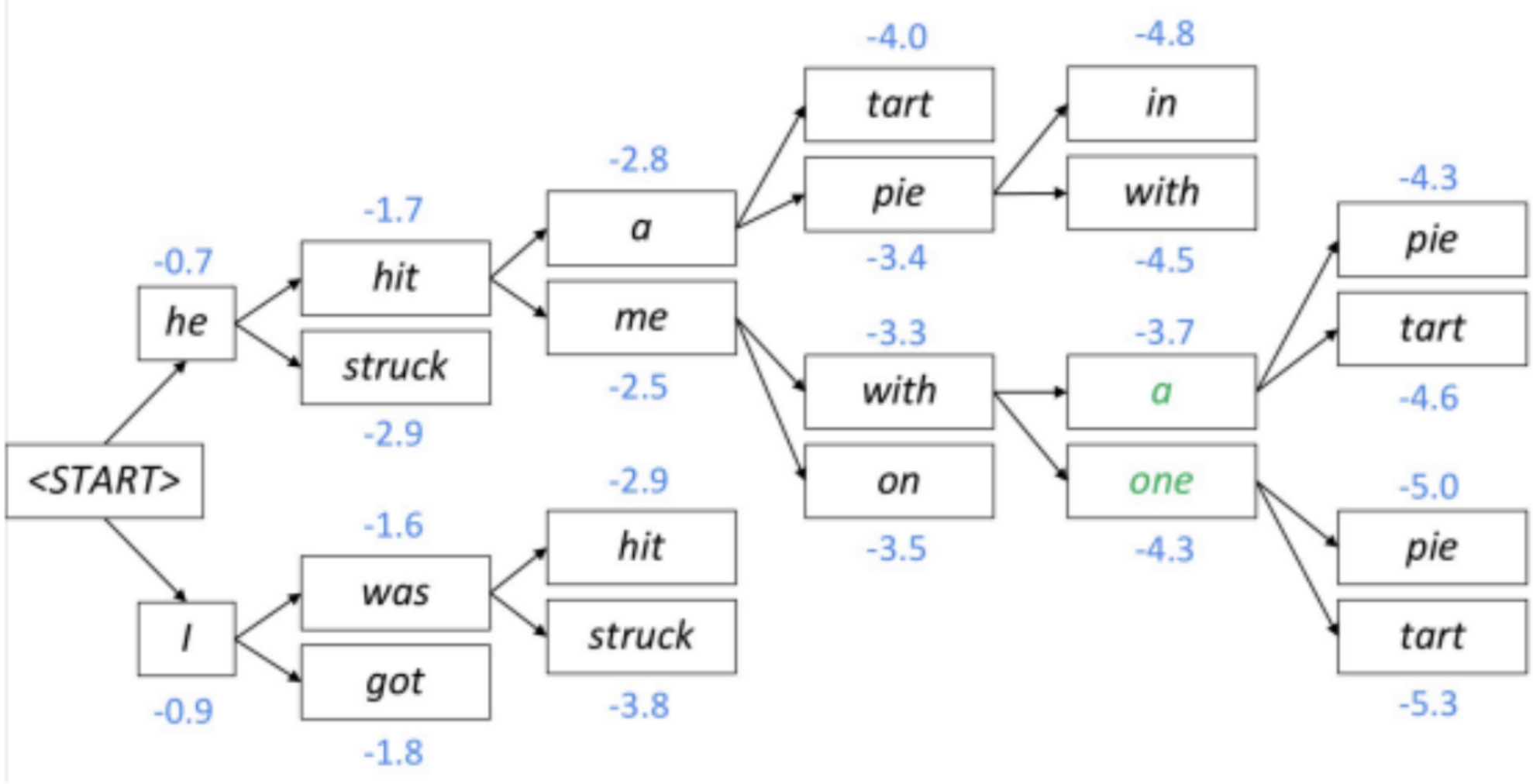

**Figure 2.6:** The process of beam search (beam size: 2) [31]

Unfortunately these sequence-to-sequence architectures have a bottleneck: the encoder cannot properly summarize the aggregated meaning of the input sequence into a singular context vector. To remedy this, $v$ is considered as a variable-length sequence, instead of a vector of a fixed dimension. On top of this with the help of the attention mechanism, we can learn the association between elements of the context vector and the output sequence.

### 2.1.5 Attention

The attention mechanism [2] was first applied to encoder-decoder models to provide a remedy to the bottleneck in plain sequence-to-sequence models, which is the limited amount of information that the context vector can encode. During each decoding step, different information could be relevant from the input sequence, so giving the decoder the same fixed-size context vector constrains performance. For this reason, an additional input is introduced at each decoding step derived directly from the encoder states. This extra input $\boldsymbol{c_i}$ is computed by taking a weighted sum over each of the encoder hidden states $\boldsymbol{h}$ as:

$$\boldsymbol{c_i} = \sum_{j=1}^{T} a_{ij} \boldsymbol{h_j} \tag{2.21}$$

where $T$ denote the number of tokens in the source sentence and hence the number of hidden states in the encoder. This process is illustrated in Figure 2.7.

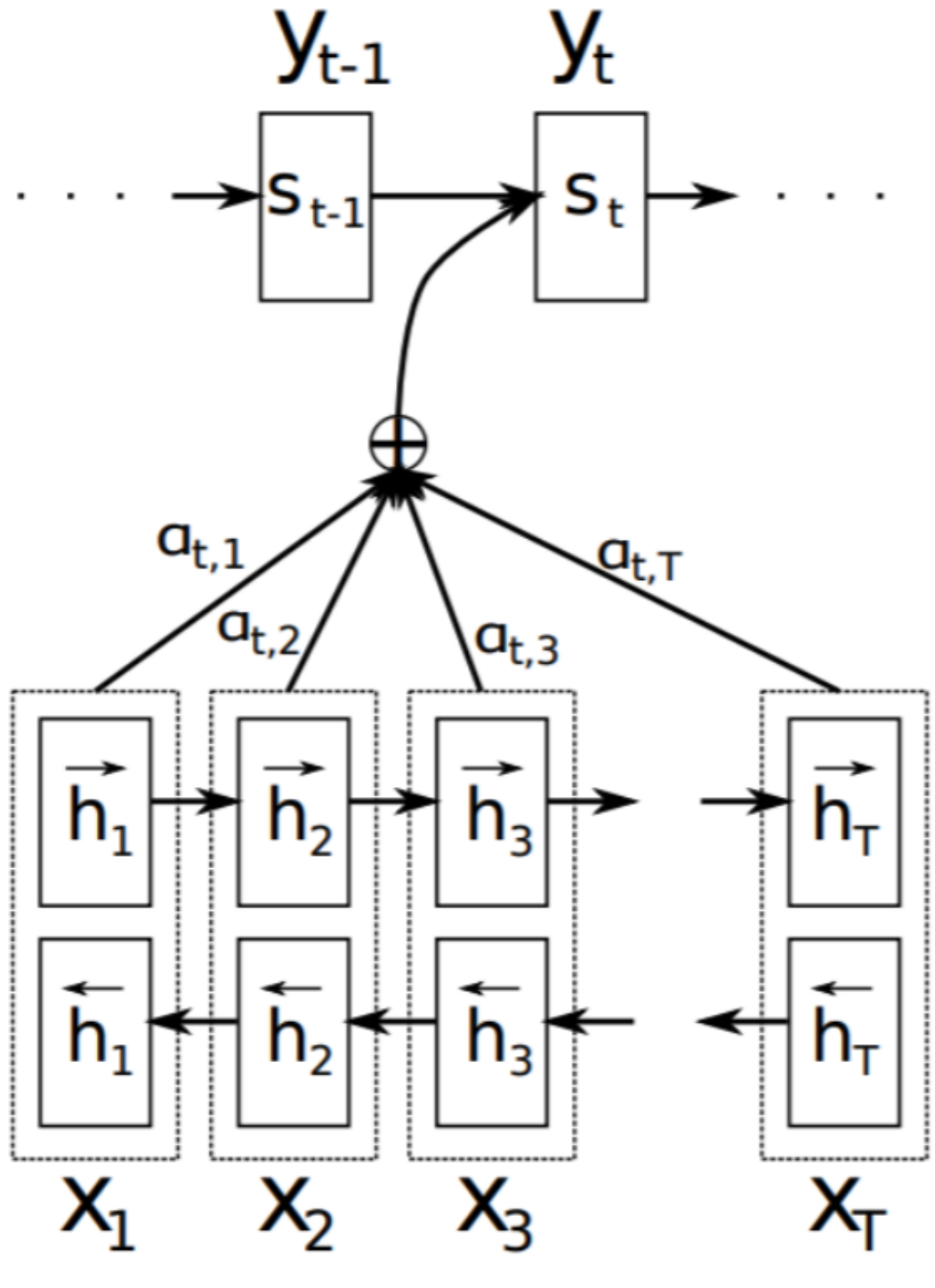

**Figure 2.7:** The attention mechanism illustrated [2]

The $a_{ij}$ weights corresponding to each hidden state $\boldsymbol{h_j}$ can be computed by a softmax function over a specific scoring function:

$$a_{ij} = \frac{exp(e_{ij})}{\sum_{k=1}^{T} exp(e_{ik})} \tag{2.22}$$

where the scoring function $e_{ij}$ is defined as

$$e_{ij} = f(\boldsymbol{s_{i-1}}, \boldsymbol{h_j}) \tag{2.23}$$

where $\boldsymbol{s_{i-1}}$ is the previous hidden state of the decode. Figure 2.8 shows a visualization of the weights $a_{ij}$ between a source and a target sentence.

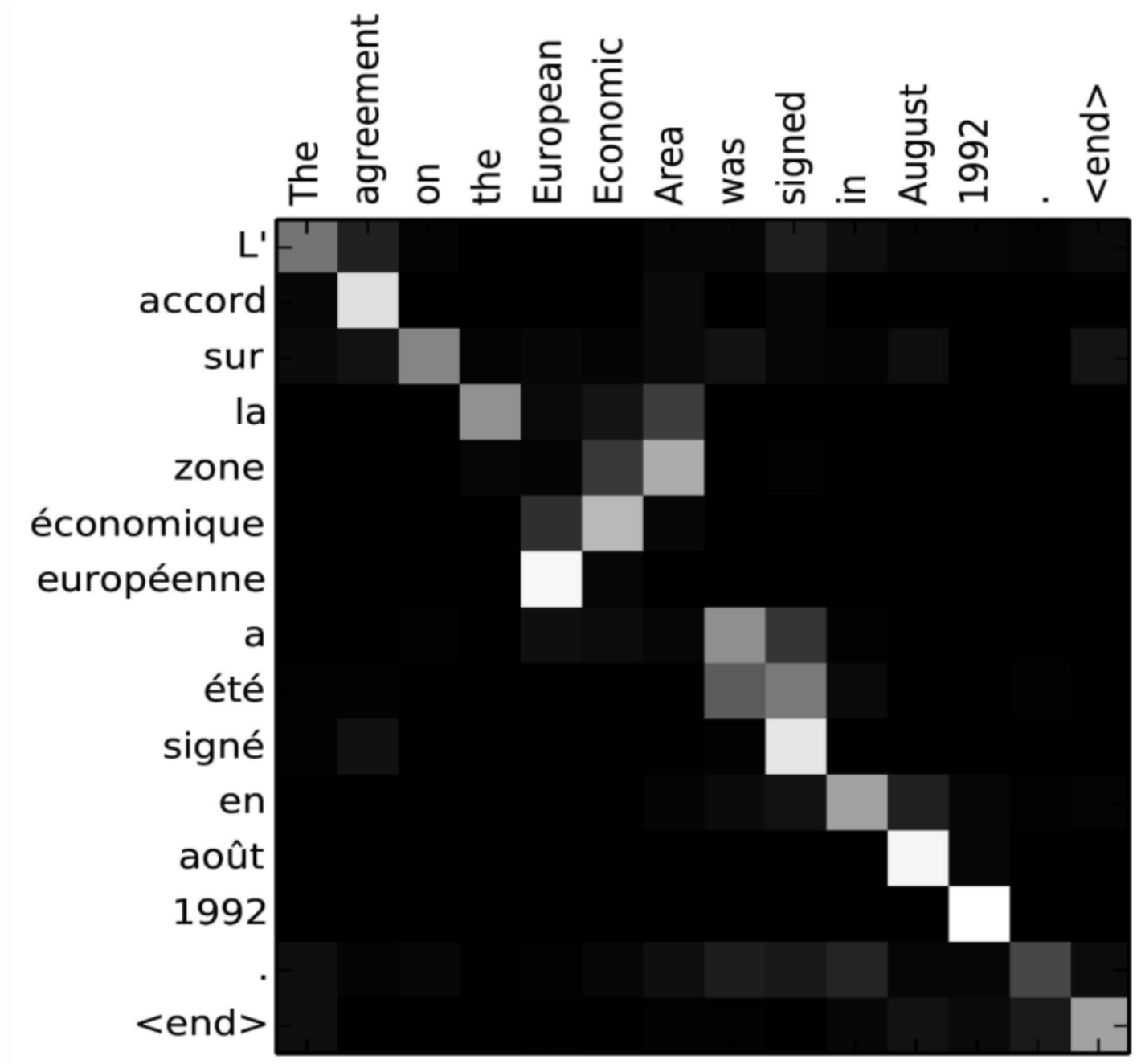

**Figure 2.8:** An attention map, showing the correlations between the source and target tokens. [2]

Using the terminology of information retrieval, $\boldsymbol{s_{i-1}}$ is often referred to as the query vector ($\boldsymbol{q}$) with $\boldsymbol{h_j}$ being called the key vector ($\boldsymbol{k}$).

The alignment score functions can be implemented in multiple ways. In the original paper, the query and key vectors are concatenated and fed into a multi-layer feed-forward neural network (Bahdanau attention) [2]:

$$MLP(\boldsymbol{q}, \boldsymbol{k}) = \boldsymbol{w_2} tanh(W_1[\boldsymbol{q} : \boldsymbol{k}]) \tag{2.24}$$

where $\boldsymbol{w_2}$ and $W_1$ are learnable parameters of the network. Luong et. al proposed [33] other variants of the attention mechanism, where the difference from Bahdanau attention is in the implementation of the alignment score function. The Dot Product (DP) scoring function is a simple matrix multiplication of the query and key vectors:

$$DP(\boldsymbol{q}, \boldsymbol{k}) = \boldsymbol{q^T k} \tag{2.25}$$

where the dimension of $\boldsymbol{q}$ and $\boldsymbol{k}$ need to be the same. The Dot Product scoring function has a bottleneck, because the scale of dot product increases as dimensions get larger. As a remedy, Vaswani et al. introduced [53] the Scaled Dot Product (SDP) scoring function, which scales by the size of the vector:

$$SDP(\boldsymbol{q}, \boldsymbol{k}) = \frac{\boldsymbol{q^T k}}{\sqrt{|\boldsymbol{k}|}} \tag{2.26}$$

Luong et. al also introduced the Bilinear (BL) scoring function [33]:

$$BL(\boldsymbol{q}, \boldsymbol{k}) = \boldsymbol{q^T} W \boldsymbol{k} \tag{2.27}$$

Attention does not necessarily need to be computed using a source and target sequence. *Self-attention* (also called intra-attention) [6] works with any of the previously defined scoring functions, with the only condition that the source and target sequence is the same. The hidden representation $\boldsymbol{h}$ in this case is a linear combination of the inputs:

$$\boldsymbol{h} = \alpha_1 \boldsymbol{x_1} + \alpha_2 \boldsymbol{x_2} + ... + \alpha_t \boldsymbol{x_t} \tag{2.28}$$

which can be rewritten as a matrix product:

$$\boldsymbol{h} = \boldsymbol{X}\boldsymbol{a} \tag{2.29}$$

where $\boldsymbol{X} \in \mathbb{R}^{n \times t}$ is the input matrix and $\boldsymbol{a} \in \mathbb{R}^n$ is a column vector containing the $\alpha_i$ attention weights. As there is only a single sentence, the query and key vectors both come from the same hidden states and therefore it is possible to create a new representation of the same sentence by relating different positions. For the same reason, self-attention can be implemented as a singular layer that takes a sentences as input and outputs a different representation. Hence it is applicable both in the encoder and the decoder. Self-attention has shown improvements for a wide variety of NLP tasks such as reading comprehension [6], abstractive summarization [41], neural machine translation [53] and representation learning [32]. Figure 2.9 shows the weights of the self-attention mechanism.

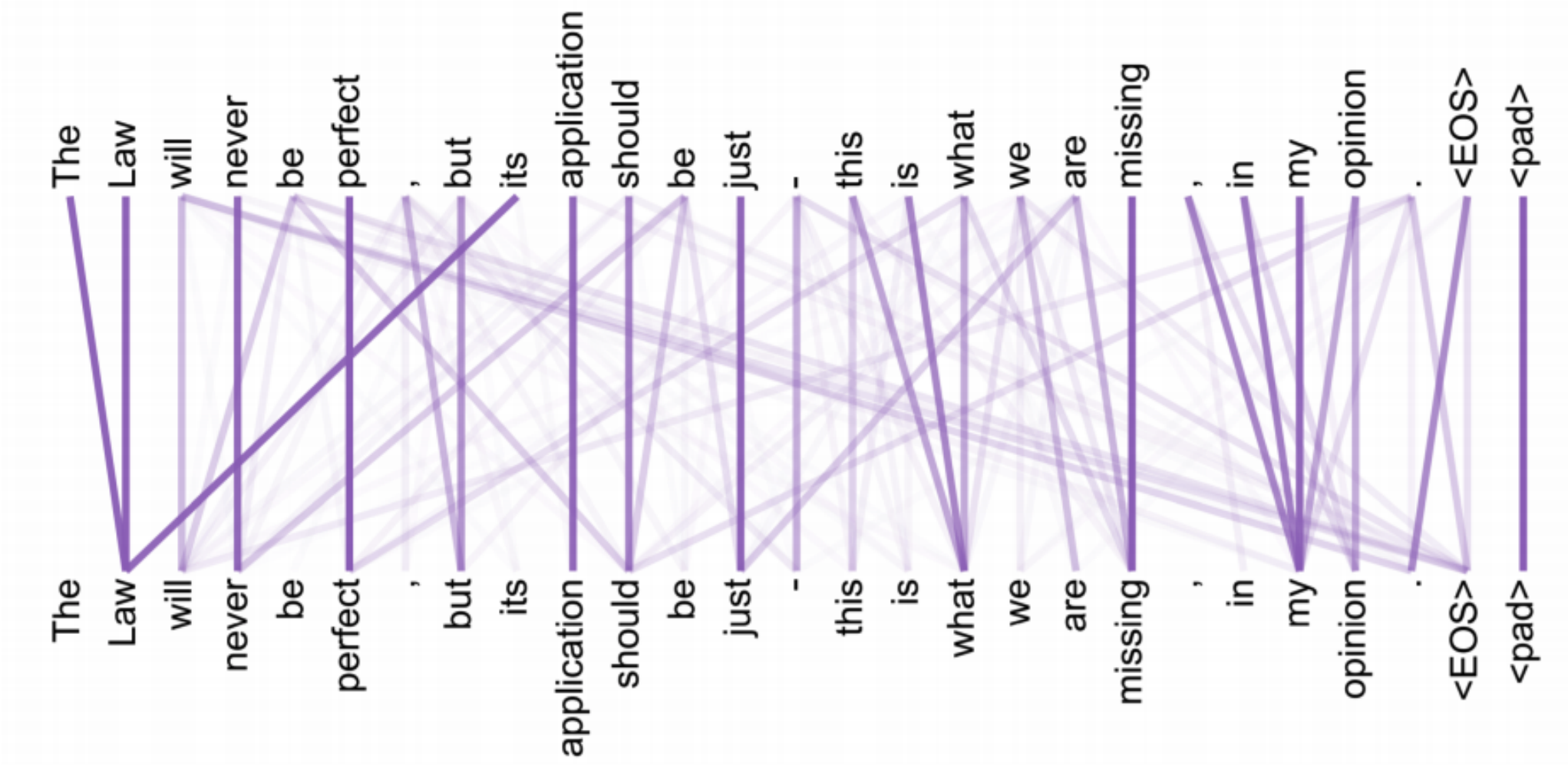

**Figure 2.9:** Self-attention visualized, where lines represent the correlation (weights) between words [53].

Regardless of the type of attention, it is possible to impose a constraint on $\boldsymbol{a}$. In case of *hard attention*, $\|\boldsymbol{a}\|_0 = 1$, which ensures that $\boldsymbol{a}$ is a one-hot vector. This way only one of the coefficients is equal to one, all the others are zero, thus the output representation is reduced to the $\boldsymbol{x}_i$ corresponding to $\alpha_i$. If we impose, that $\|\boldsymbol{a}\|_1 = 1$, the output representation is a linear combination of the inputs, where $\sum_{i=1}^{t} \alpha_i = 1$. This is called *soft attention*.

### 2.1.6 The Transformer

Although the transformer model originally achieved state-of-the-art results in machine translation [53], it surpassed previous models in a wide variety of other NLP tasks [44, 12].

In this section, I briefly discuss the architecture of the transformer, which can be seen in Figure 2.10. Transformers are composed of two main sub-components, the encoder and decoder networks. The encoder network receives the source sentence in the form of a word embedding and produces a representation of the sentence. After this, using this representation and the word embeddings of already generated tokens taken from the output sentence, the decoder produces output probabilities for each word. For this reason, the decoding process is auto-regressive, because every prediction is dependent on previously predicted words.

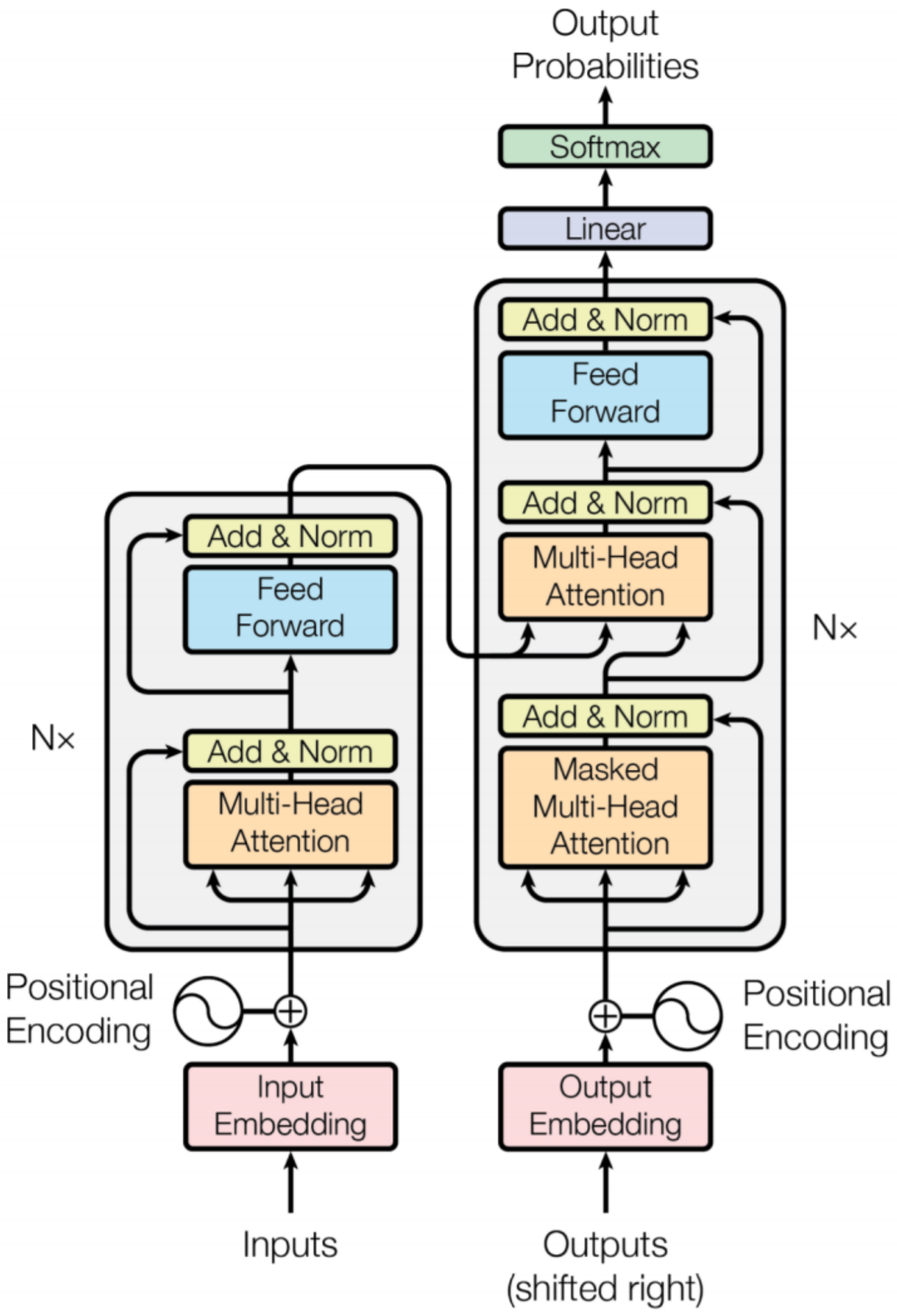

**Figure 2.10:** The architecture of the transformer, showing the encoder (left) and decoder (right) sub-networks [53].

#### 2.1.6.1 Encoder and decoder networks

The encoder of the vanilla transformer is made up of 6 homogeneous layers, that are stacked on top of each other, meaning that the output of the $n$th layer serve as the input of the $n+1$th layer. Each layer is composed of two parts: a multi-head self-attention block, and a position-wise feed-forward network. Both blocks also contain skip connections and layer-level normalization.

The architecture of the decoder is similar to the encoder as it contains 6 layers as well, although the internal blocks that make up the layers differ. An additional block is placed between the multi-head self-attention and feed forward blocks, which serve as a connection between the encoder and the decoder. This block implements a multi-head attention over the output of the encoder. As mentioned above, the first block of the decoder take the embeddings of already generated tokens from the output sequence in an auto-regressive manner. As the input size is fixed, masking is applied to guarantee that tokens, which are yet to be generated have no influence in the computation of attention. On top of this, output embeddings are shifted by one, which together with the masking guarantee that predicting token $t$ only depends on tokens generated prior to $t$.

#### 2.1.6.2 Attention blocks

Attention in transformers is instrumented as a multi-head scaled dot product over query, key and value vectors, which are respectively denoted by $\boldsymbol{q}$, $\boldsymbol{k}$ and $\boldsymbol{v}$. In case of the self-attention blocks, these vectors are constructed from the input embeddings, such that the input is projected three times with separate representations, to make up queries, keys and values of the same dimension as the input. The attention block that serves as a connection between the encoder and the decoder receive the query vectors as a projection of the output of the previous attention block in the decoder. The keys and values are constructed by projecting the output of the final encoder layer two times. As a summary, scaled dot-product attention can be formalized in the following equation:

$$Attention(Q, K, V) = softmax\left(\frac{QK^T}{\sqrt{d_k}}\right)V \tag{2.30}$$

where $d_k$ is the dimension of the key vectors and $Q$,$K$,$V$ denote matrices that are created by respectively concatenating $\boldsymbol{q}$, $\boldsymbol{k}$ and $\boldsymbol{v}$ across the input sequence.

These attentions in the transformer are multi-headed, which means that multiple attentions are computed over separate inputs in a parallel fashion. A linear projection is performed over the query, key and value matrices to reduce their dimension as:

$$d_{\text{channel}} = \frac{d_{\text{input}}}{h} \tag{2.31}$$

where $d_{input}$ denote the input dimension and $h$ denote the number of heads. The linear projection is computed by learnt weight matrices for each head and $Q$, $K$ and $V$ respectively:

$$\text{head}_i = \text{Attention}(QW_{Q_i}, KW_{K_i}, VW_{V_i}) \tag{2.32}$$

where $W_{Q_i}$, $W_{K_i}$ and $VW_{V_i}$ denote the learnable parameters. Given a $d_{input} = 512$ input sequence, by adding multiple heads, the attention does not need to be computed at once over the whole sequence. If $h = 8$, attention is computed in parallel over query, key and value vectors of dimension $\frac{d_{input}}{h} = 64$. Once all channels are computed, the results need to be concatenated and once again projected to form the output of the attention block:

$$\text{MultiHeadAttention} = \text{Concatenate}(\text{head}_i, ..., \text{head}_h)W_O \tag{2.33}$$

where $W_O$ denote the learnt projection matrix corresponding to the output of the block. The scaled dot-product attention and the multi-head attention is shown on Figure 2.11.

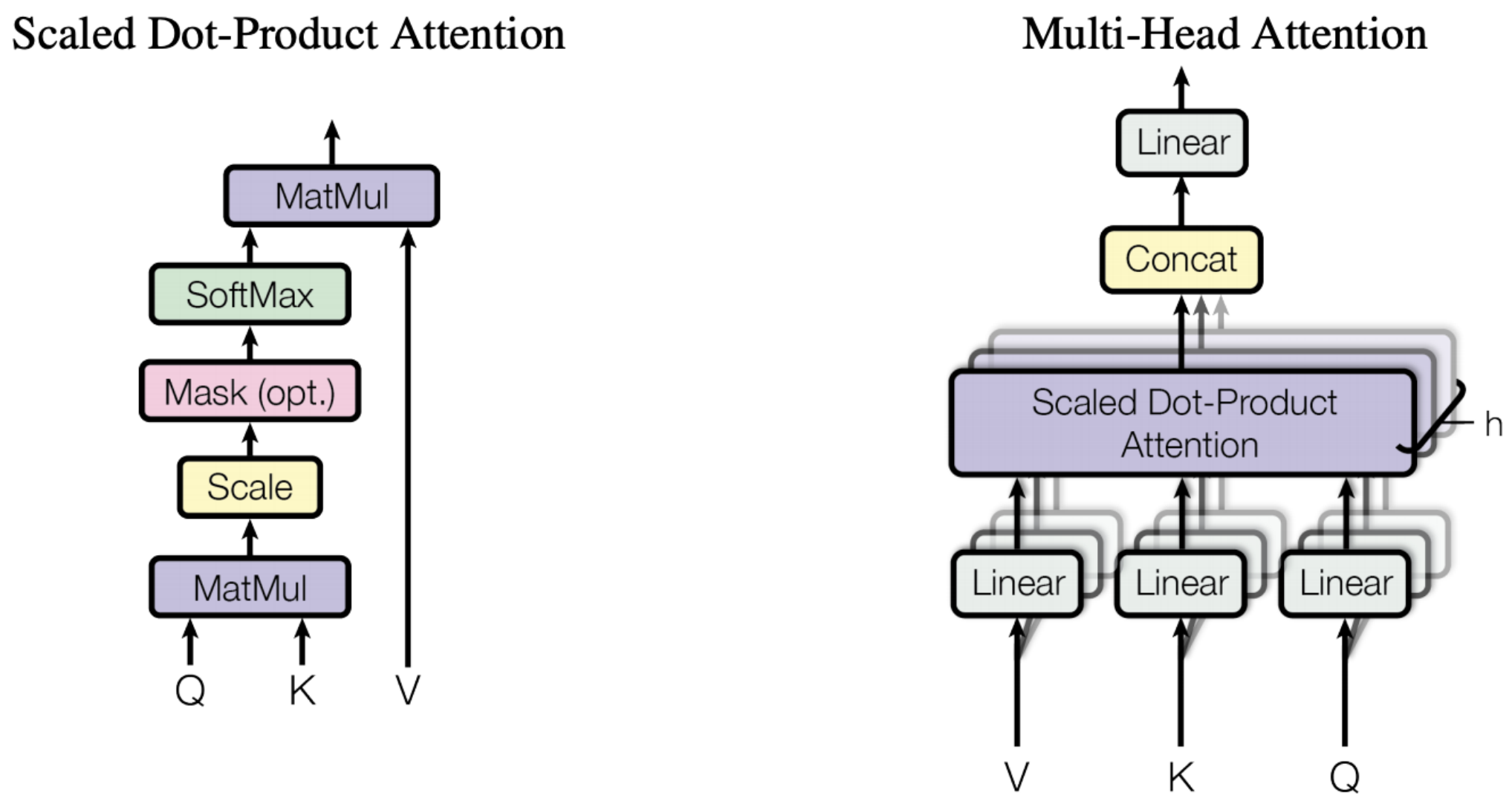

**Figure 2.11:** Scaled dot-product attention (left) and multi-head attention (right) [53], with $Q$, $K$ and $V$ respectively denoting the query, key and value matrices and $h$ being the number of attention heads.

#### 2.1.6.3 Feed-forward blocks

The feed-forward block (FFN) is composed of two linear transformations with a ReLU activation in between, which are the same for each position, although they use different parameters in every layer. Intuitively, this process can be thought of as performing two convolutions with a kernel size 1: every filter contains a single learnable parameter, which is multiplied with the input vector in an element-wise manner to produce the output vector as

$$\mathrm{FFN}(\boldsymbol{x}) = \max(0, \boldsymbol{x}W_1 + \boldsymbol{b}_1)W_2 + \boldsymbol{b}_2 \tag{2.34}$$

where $\boldsymbol{x}$ denote the input vector, $W_1$ and $W_2$ are learnable parameters and $\boldsymbol{b}_1$ and $\boldsymbol{b}_2$ are bias vectors.

#### 2.1.6.4 Regularization

Several types of regularization techniques are applied in the transformer to improve its robustness. First of all, the outputs of each layer are normalized using simple statistical metrics such as mean and variance computed on a layer level. This process is called layer normalization [1]. Furthermore, this normalization is employed along residual connections [18] between the inputs and outputs of a layer. This means, that the final output of each layer $L$ is given by the term:

$$\mathrm{LayerNorm}(X, L(X)) \tag{2.35}$$

where $X$ is the input and $L$ is function implemented by the layer. Furthermore, dropout is applied as a regularization technique [48], which is a stochastic mechanism, that replaces elements of its input with 0 with a predefined probability. Within the transformer, it is applied both at the output of each layer before layer normalization and also during the computation of the scaled dot-product.

#### 2.1.6.5 Positional encoding

Capturing positional information is an essential part of solving sequential modeling problems. In case of languages, position and word order are what primarily define grammar. As of this point, the architecture of the transformer model is solely based on attention and feed-forward networks, therefore sequential information on the input sentences are completely disregarded. The remedy to this is positional encoding, which means that for each word a separate representation is derived based on their position in the sentence. There are multiple ways to implement this positional encoding, either in a learnt or fixed manner [14]. In the original transformer, Vaswani et. al use sine and cosine functions of different frequencies:

$$\mathrm{PE}_{(pos_k,2i)} = \sin\left(\frac{pos_k}{10000^{\frac{2i}{d_model}}}\right) \tag{2.36}$$

$$\mathrm{PE}_{(pos_k,2i+1)} = \cos\left(\frac{pos_k}{10000^{\frac{2i}{d_model}}}\right) \tag{2.37}$$

where $i \in [0, 1, ..., d_{model} - 1]$ and $\mathrm{PE}_{(pos_k)}$ correspond to the positional encoding of the word at position $k$ in the sequence. It is important, that the dimension of this positional encoding vector is the same as the input vector, because the final representation of each word is computed by taking the sum of the two vectors.

## 2.2 Neural machine translation

Machine Translation (MT) is a large subfield of natural language processing, that aims to automatically translate corpora from one language to an other. It is an inherently complex task, because it involves solving and developing a general understanding to several challenges of NLP, e.g. word order, word sense ambiguity, tenses or dealing with idioms. The degree of how hard it is to create a model that carry all these linguistic capabilities vary a lot from one language-pair to the other, making each language a challenge on its own. Early approaches to machine translation relied on deriving translation rules based on our knowledge of linguistics. This, however, could not cover the many exceptions and irregularities that languages contain. As parallel corpora has become increasingly available, data-driven approaches gained dominance over rule-based methods. The predecessor of neural approaches is Statistical Machine Translation (SMT) [4, 27], which use statistical models - with parameters that are derived from the analysis of bilingual corpora - to learn latent structures in the data. Although SMT performs significantly better than rule-based approaches, it struggles with learning long-distances dependencies, which is a necessity for high quality translations. In the last decade, the growingly increasing computing power and the vast amount of available data gave ground to deep learning and with that a new paradigm has emerged: Neural Machine Translation (NMT) [22, 8, 49, 2]. It is fundamentally different from previous approaches for several reasons. Firstly, the discrete representations of prior methods are replaced with continuous representations due to the nature of neural networks. Secondly, using a single neural network to model translation makes the process completely end to end and therefore the need for tuning and exhaustive feature engineering in separate subcomponents of the system is eliminated. For these reasons, neural machine translation surpassed its predecessors and became the foundation of modern translation systems.

### 2.2.1 Overview

Machine translation can be defined on several levels of the written language, including document-, paragraph- and sentence-level. As my work is focused on sentence-level translations, I will discuss the nomenclature accordingly.

Given a bilingual dataset $\mathcal{D}$ containing sentences from the source and target language, we take the representation of such sentence pairs $\boldsymbol{x} \in \mathcal{D}$ and $\boldsymbol{y} \in \mathcal{D}$ and formulate the goal of neural machine translation as estimating the unknown conditional probability $P(\boldsymbol{y}|\boldsymbol{x})$. As these sentences are essentially just sequences of words, the formalism of sequence-to-sequence models introduced in section 2.1.4 is applicable. For this reason, the vast majority of NMT models are based on the encoder-decoder architecture, which in the simplest case consists of an embedding layer, the encoder, the decoder and a classification layer that computes the output. This architecture is shown on Figure 2.12.

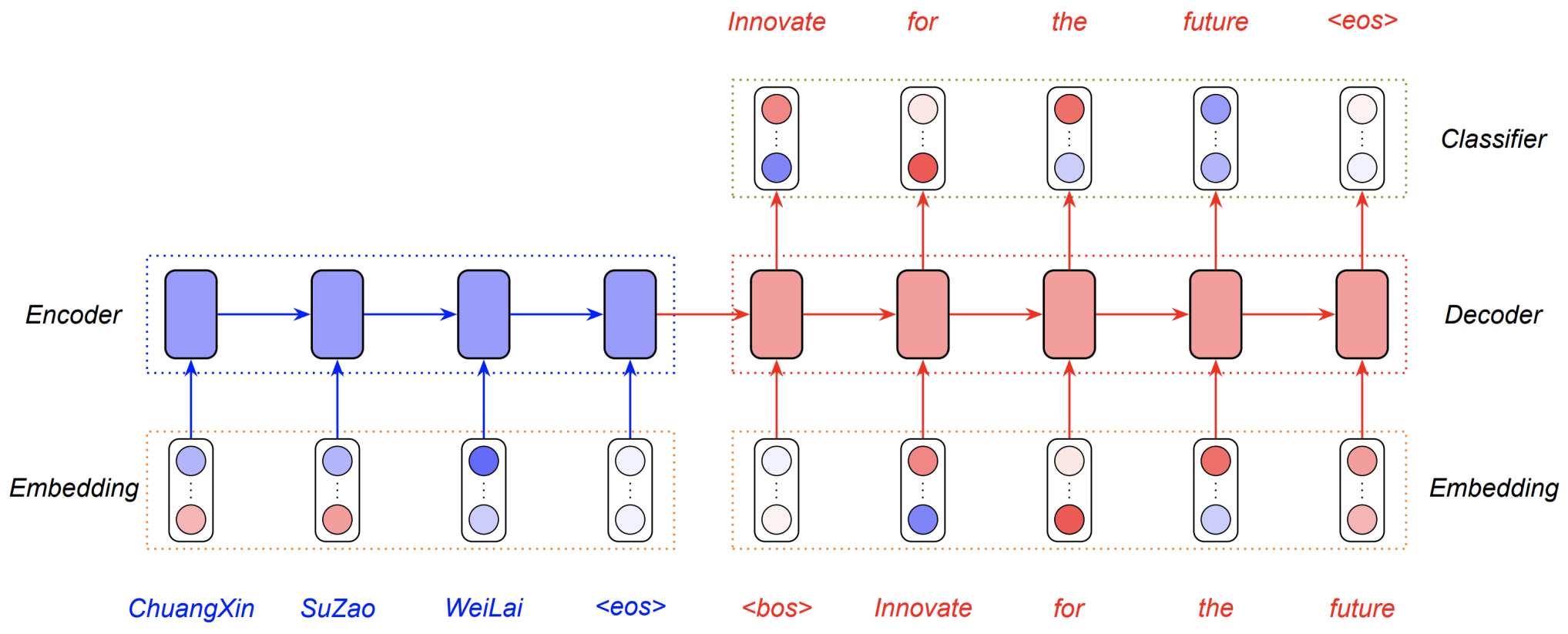

**Figure 2.12:** A simple autoregressive encoder-decoder architecture for neural machine translation [50].

The embedding layer is responsible for taking the discrete symbols $x_t$ and mapping them to a continuous representation $\boldsymbol{x_t} \in \mathbb{R}^d$, where $d$ is dimension of the representation. This then is fed into the encoder, which creates a hidden representation of these embeddings with the purpose of encoding important and expressive information about the source sentence. The decoder is responsible for extracting relevant information from the hidden representations and thus it is a language model conditioned on the output of the encoder and the already generated symbols. The classifier situated at the end of the architecture predicts the distribution of tokens at each timestep. The objective during the training of these architectures is typically the maximum log-likelihood (MLE), which means the maximization of the loss function $\mathcal{L}(\boldsymbol{\theta})$ (defined in Equation 2.19):

$$\hat{\boldsymbol{\theta}}_{\text{MLE}} = \arg\max_{\boldsymbol{\theta}} \Big\{ \mathcal{L}(\boldsymbol{\theta}) \Big\} \tag{2.38}$$

During training, the error can easily be propagated backwards in the entire architecture, allowing the computation of the gradient of $\mathcal{L}$ with respect to $\boldsymbol{\theta}$. Using this gradient, the parameter updates can be done with any optimizer, such as SGD (Equation 2.6) or an adaptive learning rate optimizer, like Adam [24].

### 2.2.2 Architectures

Having formulated the problem of machine translation as a sequence to sequence problem, it is worth diving into the relevant architectures that have been impactful during the last decade. As neural machine translation has always been a central problem in sequence modeling and NLP, the dominant architectures in NMT closely followed the development of sequential modeling and also experienced the drawbacks of each architecture. Early neural models employed a fixed-length representation of the source sentence and used recurrent neural networks during decoding to create translations of variable length [22, 8]. Cho et. al showed that as the source sequence length increases, the fixed-length representation on the source side is becoming a bottleneck in the system, because the encoder is not capable of compressing enough information about the source sentence into a single vector [7]. Furthermore, capturing long-term dependencies has been a well-known challenge in sequential modeling [3]. In this regard, a relevant indicator that needs to be taken into account is the longest path between source and target tokens. With fixed-length representations this path is $O(S+T)$, which makes capturing the long-term dependencies difficult. A key turning point in the development of architectures was the introduction of the attention mechanism (see Section 2.1.5), which reduces the path between source and target words to a constant length. Sequence to sequence models originally were composed of recurrent neural networks, however it is possible to use different architectures, such as convolutional neural networks as well [23, 14]. The real breakthrough however was the transformer (see Section 2.1.6), which became the basis of the current state of the art translation models.

## 2.3 Dependency Parsing

The field of natural language processing is dominated by supervised learning and neural networks in particular, that represent the meaning of words in a high-dimensional vector space. The issue in many practical applications of NLP and also machine learning in general is that there is very limited visibility into the decision-making of neural networks and also the high dimensional representations, upon which these decisions are made. Contrary to this approach, there are methods that intend to reveal the connections in the data and using this information create an explicit representation. Dependency parsing is a such method, which captures syntactic relations in a sentence by representing the grammatical relations between words as a directed acyclical graph. The syntactical structure of this graph is composed of lexical elements, among which a set of binary relations are defined. The elements of these binary relations are usually referred to as head and dependent. An example dependency tree for English and Hungarian are shown on Figures 2.13 and 2.14.

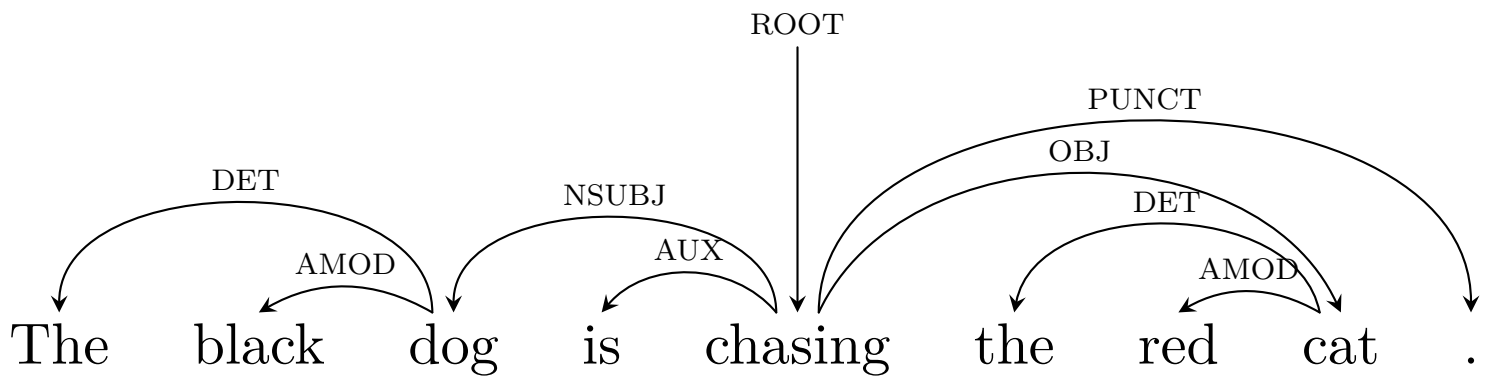

**Figure 2.13:** An example dependency parse tree for English.

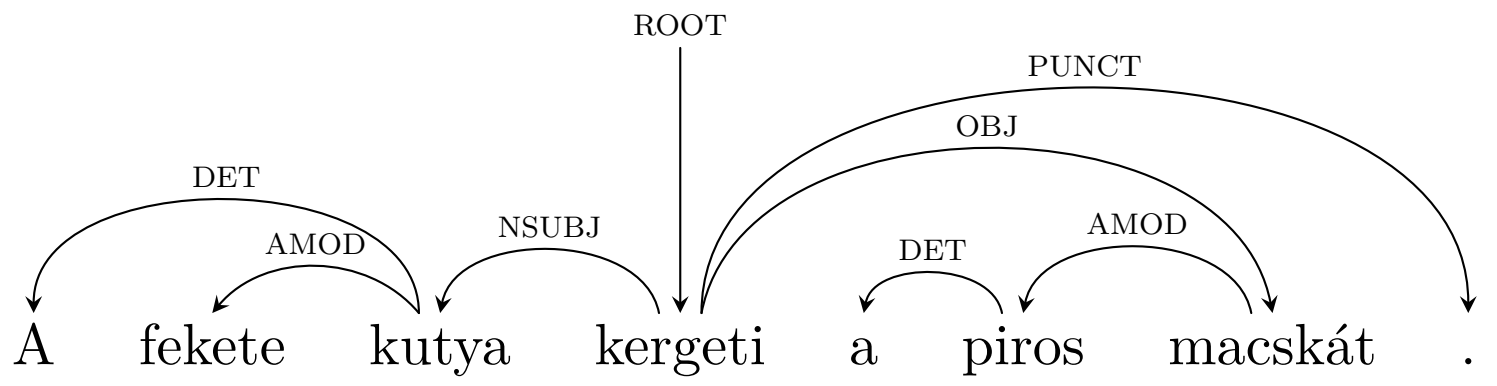

**Figure 2.14:** An example dependency parse tree for Hungarian.

Several formalisms exist that define the possible semantics that can exhibit in these relations, which many times differ from language to language. There is an ongoing and common effort to create a language-independent framework for consistent grammatical annotations: Universal Dependencies[1] (UD) [11]. The goal of UD is to define a universal taxonomy of categories and annotation guidelines, with the possibility of adding language-dependent extensions. It defines two main types of core dependencies. Clausal relations represent syntactic roles with respect to the predicate of the sentence, while modifier relations define how a dependent modifies their heads [10]. Table 2.1 summarizes some of the common labels in Universal Dependencies.

| **Clausal Argument Relations** | **Description** |
|---|---|
| NSUBJ | Nominal subject |
| DOBJ | Direct object |
| IOBJ | Indirect object |
| CCOMP | Clausal complement |
| XCOMP | Open clausal complement |
| **Nominal Modifier Relations** | **Description** |
| NMOD | Nominal modifier |
| AMOD | Adjectival modifier |
| NUMMOD | Numeric modifier |
| APPOS | Appositional modifier |
| DET | Determiner |
| CASE | Prepositions, postpositions and other case markers |
| **Other Notable Relations** | **Description** |
| CONJ | Conjunct |
| CC | Coordinating conjunction |

**Table 2.1:** Selected dependency relations from the Universal Dependencies [10].

As my work involves generating dependency trees for multiple languages, UD is a well-suited framework for my experiments. Luckily there are out of the box NLP pipelines, that support dependency parsing, so there is no need to implement a dependency parser from scratch. For English, I use the Stanza pipeline[2] developed by the Stanford NLP group [43]. See Figure 2.15 for a high-level overview of the pipeline. For Hungarian dependency parsing, I used the Hungarian Spacy models[3].

[1] https://universaldependencies.org/
[2] https://stanfordnlp.github.io/stanza/index.html
[3] https://github.com/oroszgy/spacy-hungarian-models

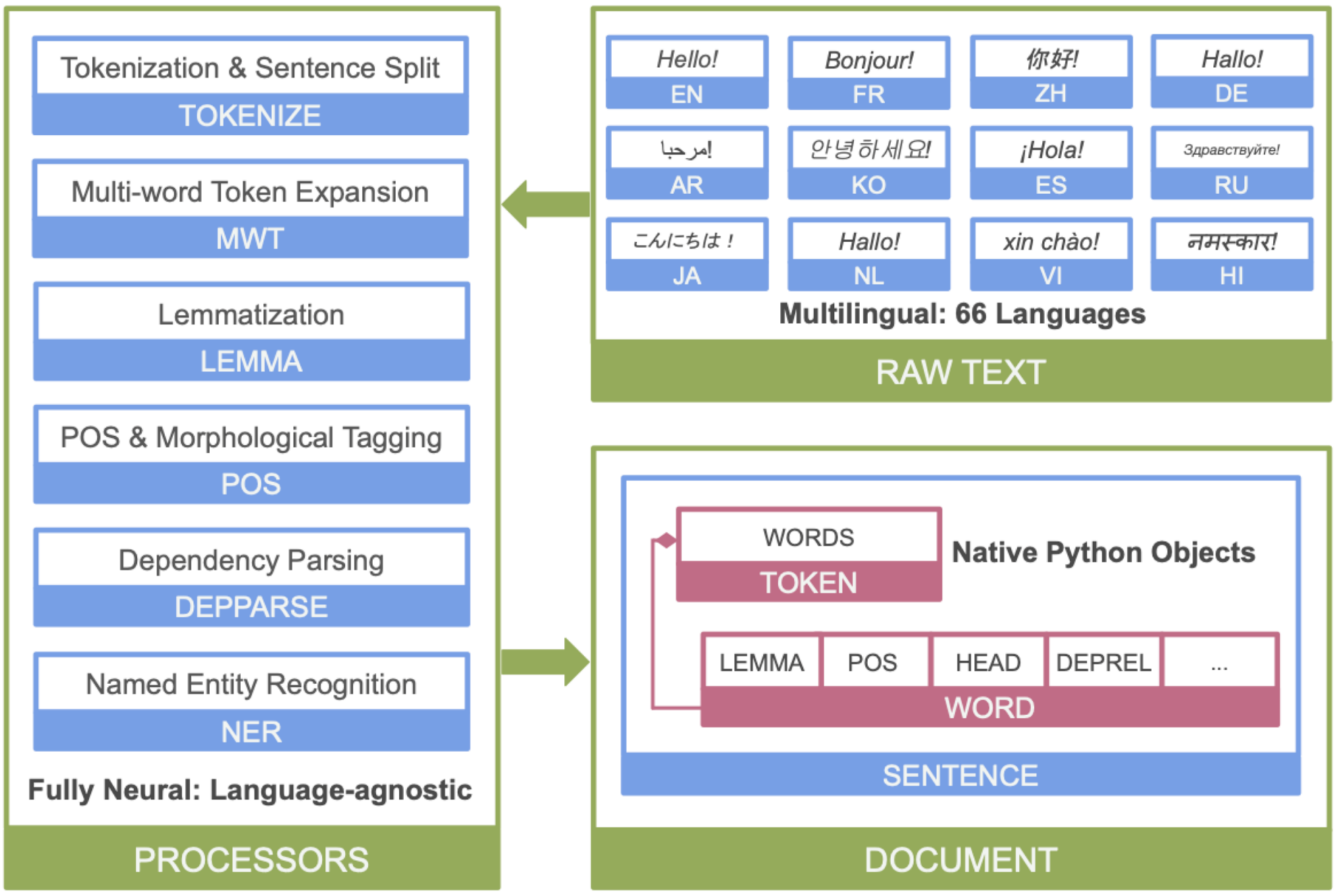

**Figure 2.15:** High-level overview of the Stanza NLP pipeline [43].

## 2.4 Data augmentation

Deep learning methods perform remarkably well on a wide variety of tasks. The superior performance of these models rely heavily on the quantity and quality of training data, because with small datasets models are likely to overfit and generalize poorly. Collecting and annotating new, good quality datasets is a wearing and time consuming task, both in industrial applications and the academia. For this reason, approaches that eliminate the need for large amounts of annotated data are gaining more and more attention (e.g. unsupervised or semi-supervised learning). Another way to tackle this problem is automatically generating new data points by augmenting existing ones with a predefined algorithm. This is a generic approach that works well in supervised learning and can be as simple as cropping/rotating images in case of image classification [38] or more complex such as creating synthetic data points with generative adversarial networks [16]. Krizhevsky et al. showed that increasing the size of the training set through augmentation contributed to the reduction of overfitting and increased model performance [28]. Although data augmentation methods gained popularity primarily on image classification, they are similarly applicable to textual data. For multi-class classification, Wang et al. stochastically select words from a sentence for replacement, such that the new word is close to the old one in an embedding space according to a distance metric (such as cosine distance) [54]. Mueller et al. replaced random words and trained siamese neural networks for sentence similarity [36]. Kobayashi trained a language model to predict a word based on its surrounding context and used this model to replace words in a sentence with the purpose of augmentation [26]. Xie et al. inject noise in the training data by randomly replacing words with a blank token [56], which, helps the model to not overfit to specific contexts. As translation models are very data-intensive to train, data augmentation methods provide particularly high value. Given a sentence $s$ in the source language, the shared encoder is used to transform $s$ to a latent represenation, which is then fed into the decoder of the target language, creating a noisy translation $\hat{t}$. It is possible to predict the original sentence $s$ using this noisy translation and the encoder and decoder of the source language (Figure 2.16).

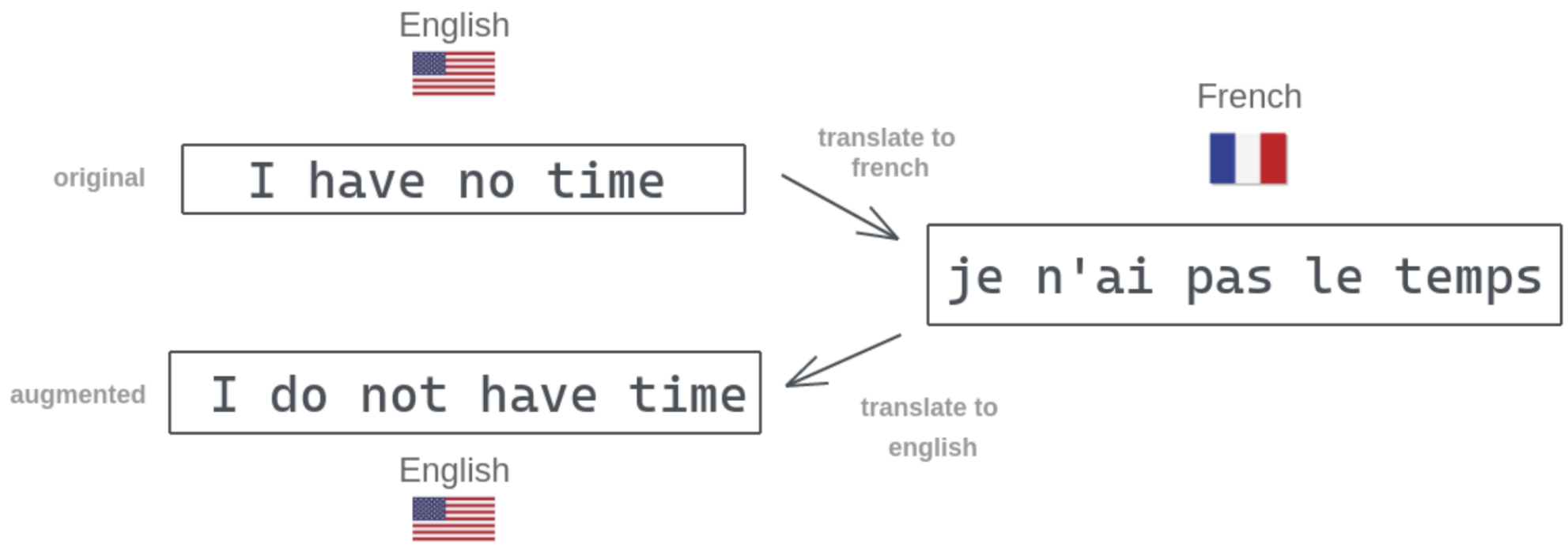

**Figure 2.16:** The process of backtranslation [5]

This method is called backtranslation [47] and it has proven to be an effective data augmentation method in NMT, because we can create a pseudo-parallel corpus, which serve as useful training data, particularly for low-resource languages.

The majority of the augmentation methods described above build upon the hypothesis, that adding an appropriate amount of noisy data in the training set results in more robust models and better generalizations. The way these noisy samples are generated often involves random selection of words, not taking into account syntax, semantics and based on these: the importance of words in a sentence. As seen in Section 2.3, dependency parsing provides insight into the semantic relations among words and therefore can serve as a basis for augmentations that make more sense semantically and have less grammatical errors. Several studies have shown that that such structure-aware augmentation methods are effective in a wide variety of tasks. Xu et al. achieve improvements on word relation classification using dependency path based augmentation [57]. Sahin and Steedman show that performing swapping and rotation on subtrees of a dependency parse tree can benefit POS tagging for low-resource languages [46]. Vania et al. build further evaluations on top of the work of Sahin and Steedman and conclude that augmentation via dependency tree morphing is beneficial for the task of dependency parsing. Duan et al. use the depth of dependency trees as a clue for the importance of words when performing simple augmentation methods [13]. They apply this method on neural machine translation, such that the source sentence is transformed by augmentations operators and the target sentence is left intact. The augmentation operations applied are the following:

- **Blanking** [56]. Selected words are replaced with a placeholder token (e.g. BLANK).
- **Dropout** [21]. Selected words are removed from the sentence.
- **Replacement** [56]. Selected words are replaced with a word that has a similar unigram word frequency over the corpus.

| Original | We | shall | fight | on | the | beaches | . |
|---|---|---|---|---|---|---|---|
| Select or Not | no | no | no | yes | no | yes | no |
| Blanking | We | shall | fight | BLANK | the | BLANK | . |
| Dropout | We | shall | fight | | the | | . |
| Replacement | We | shall | fight | with | the | sandy | . |

**Table 2.2:** An example of the augmentations used by Duan et al [13]

Duan et al. hypothesize that the meaning of a sentence is determined only by a handful of words, which are likely to be situated closer to the root node of the dependency tree. For this reason, instead of randomly selecting words for augmentation, they calculate the $q_i$ chance of selecting a word based on its depth in the dependency tree:

$$q_i = 1 - \frac{1}{2^{d^i - 1}} \tag{2.39}$$

where $d_i$ denote the tree depth. The final probability distribution for selecting a word for augmentation is computed by a softmax function.

# Chapter 3

# Methods

## 3.1 Structure-aware data augmentation

The main contribution of my thesis is developing new structure-aware data augmentation techniques. In this section, I will briefly discuss these techniques and outline the algorithms that I created for data augmentation.

When performing data augmentation for NMT, it is often an issue, that augmentation is only performed on the source sentence while the target sentence is left untouched. This is reflected in the work of Duan et al. as well [13], because although noise can improve robustness, there is a chance of distorting the meaning of the sentence to a degree, where the translation is not valid anymore. As an example, taking the *Replacement* method from Table 2.2, it is clearly visible how this approach can lead to false translations (Table 3.1).

| | |
|---|---|
| Original-EN | We shall fight on the beaches. |
| Original-HU | A tengerparton kellene küzdenünk. |
| Replacement | We shall fight with the sandy. |
| Replacement-HU (Google Translate) | Harcolni fogunk a homokkal. |

**Table 3.1:** Error analysis of the *Replacement* data augmentation method [13]

The motivation of this work is to develop augmentation methods for NMT, which perform the augmentation simultaneously on both the source and the target sentence. The way this is achieved is by computing the dependency trees for both the source- and target-side sentences and performing predefined augmentations simultaneously based on the dependency parse trees. This way we hypothesize, that new and useful training samples can be created, while also keeping the translations more accurate. The augmentation algo-rithms are based on finding a specific substructure within the dependency tree, containing a triplet of the subjects, objects and predicates of sentences (as seen in Figure 3.1). The different augmentation methods are based on swapping parts of this substructure of the dependency trees within sentences and then creating new sentences by reconstructing the the augmented dependency trees.

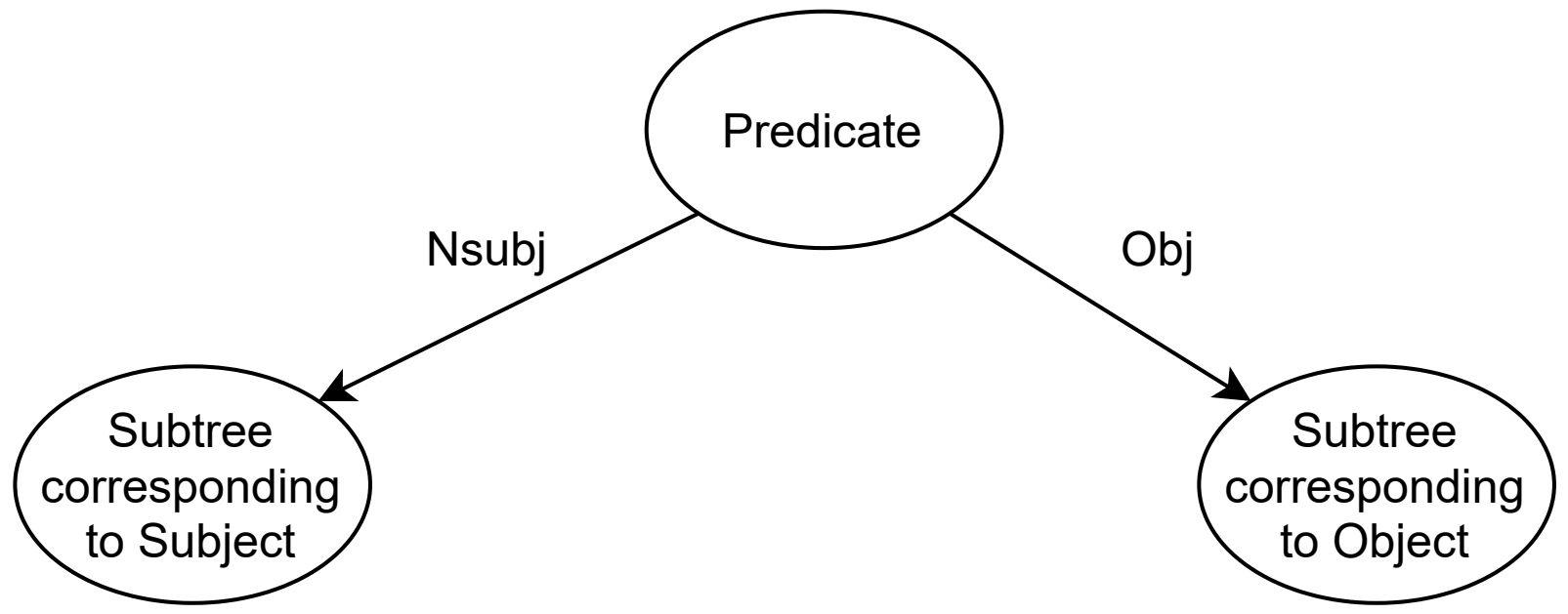

**Figure 3.1:** The substructure of subject, object and predicate in a dependency graph.

### 3.1.1 Data filtering

The substructure seen above is very common in most languages, including Hungarian and English. Given a dataset $\mathcal{D}$, which has a subset $N \in \mathcal{D}$ that contains the desired substructure, the subtree swappings can be performed between any two of the sentences vice versa, meaning that for one augmentation method we can generate $N * (N - 1)$ sentences. For example, if $N = 10^6$ and we only perform one kind of swapping, it is possible to generate $9.99999 * 10^{11}$ number of augmented sentences.

If we apply multiple methods individually or start swapping multiple subtrees at the same time, the problem space easily explodes. For this reason, it makes sense to apply data filtering on the dataset based on both the original sentences and their respective dependency graph representations. Computing dependency trees for millions of sentences is time consuming, so removing sentences that are likely to be not eligible for augmentation is important on the sentence level. To achieve this, filtering is applied based on token length of the source and target sentences respectively and also by the ratio of the source-target token lengths. Having computed the dependency trees for eligible sentence-pairs, we constrain that the following conditions need to be satisfied:

- The dependency trees contain only one subject and object (one *Nsubj* and one *Obj* edge) for both the source and target sentences.

- The subtrees corrseponding to objects and subjects contain a consecutive sequence of words with respect to the original word order (see Figure 3.2).

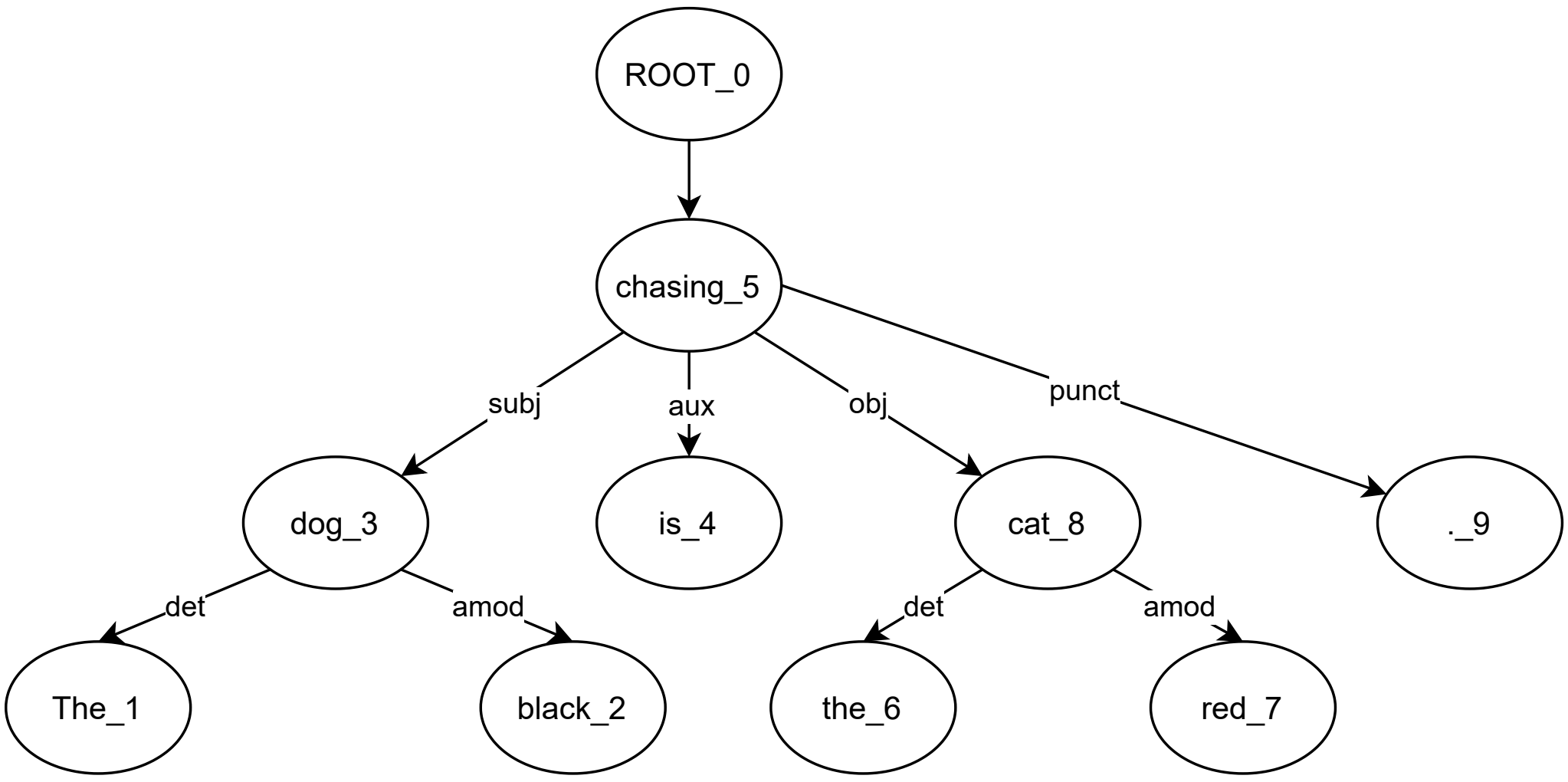

**Figure 3.2:** This figure shows the dependency parse tree of the sentence: "The black dog is chasing the red cat." This tree is eligible for augmentation, because the subtree corresponding to the subject (dog_3, the_1, black_2) form a consecutive subsequence if the original word order is reconstructed. The same is true for the subtree of the object.

### 3.1.2 Algorithms

Having found a subset of the sentence-pairs that are eligible for augmentation, in this section I describe the algorithms used for creating new sentences and also illustrate them with examples. Each method iteratively samples two sentence pairs from the training dataset without replacement and performs specific augmentation operations, that are defined below. Figure 3.3 illustrates the three main substructures that are swapped during augmentations: the subject, the object and the predicate. The below example highlights the corresponding substructures in a dependency tree pair for Hungarian, although it is important to note that we perform these swaps in parallel for both the source and target sentences.

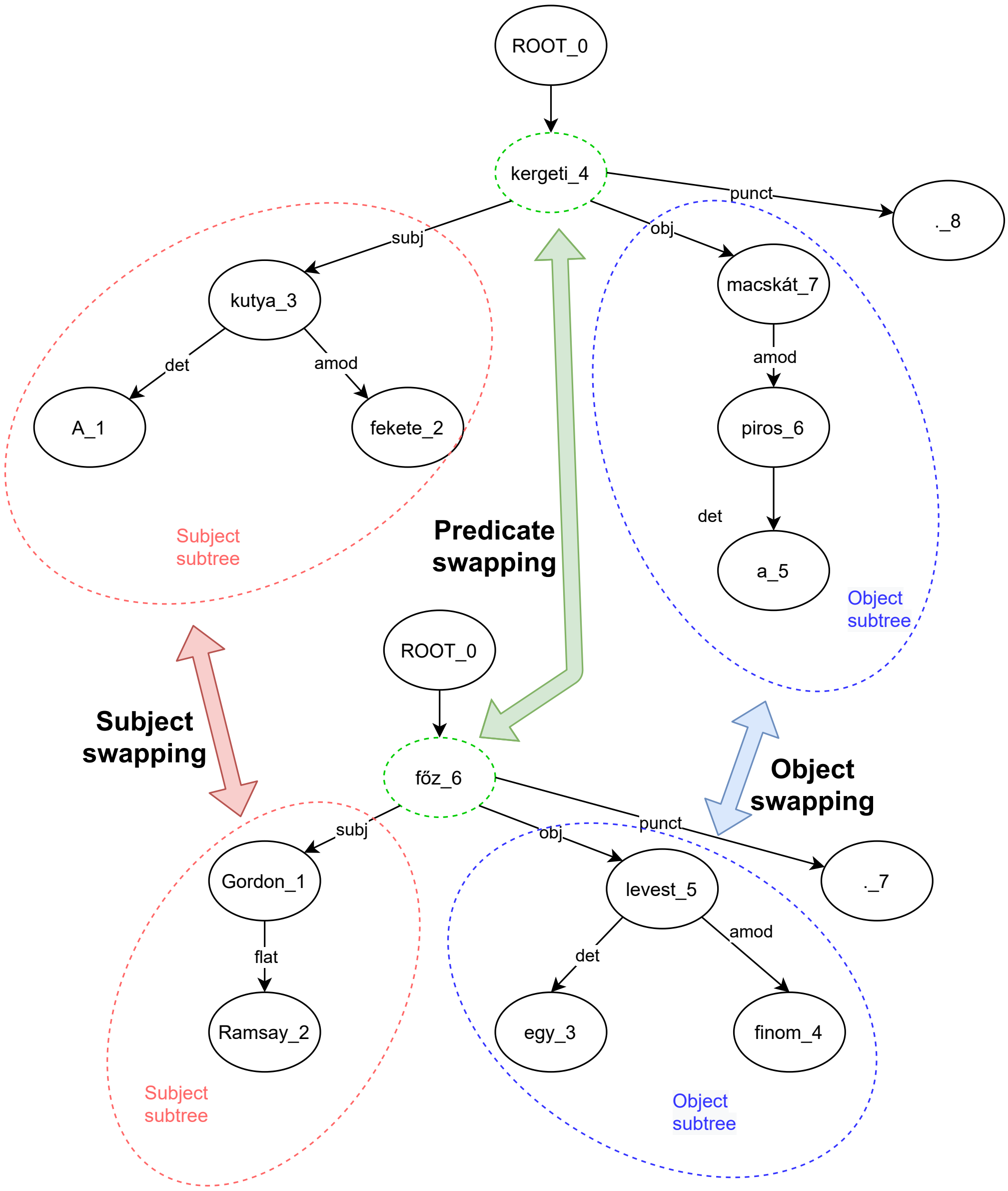

**Figure 3.3:** Overview of the proposed data augmentation algorithms on a Hungarian sentence pair.

#### 3.1.2.1 Object swapping

Given two sentence pairs of source and target sentences, we simultaneously swap the subtrees corresponding to the object of the sentence. An example augmentation and the generated sentences are shown in Table 3.2.

| | |
|---|---|
| Sentence1-EN | The black dog is chasing the red cat. |
| Sentence1-HU | A fekete kutya kergeti a piros macskát. |
| Sentence2-EN | Gordon Ramsay is cooking a delicious soup. |
| Sentence2-HU | Gordon Ramsay egy finom levest főz. |
| EN-AUG-1 | The black dog is chasing a delicious soup. |
| HU-AUG-1 | A fekete kutya egy finom levest kerget. |
| EN-AUG-2 | Gordon Ramsay is cooking the red cat. |
| HU-AUG-2 | Gordon Ramsay a piros macskát főz. |

**Table 3.2:** Augmentation via subtree swapping of objects.

#### 3.1.2.2 Subject swapping

Given two sentence pairs of source and target sentences, we simultaneously swap the subtrees corresponding to the subject of the sentence. An example augmentation and the generated sentences are shown in Table 3.3.

| | |
|---|---|
| Sentence1-EN | Sauron has regained much of his former strength. |
| Sentence1-HU | Szauron szinte teljesen visszanyerte az erejét. |
| Sentence2-EN | A hooded figure has followed us into the woods. |
| Sentence2-HU | Egy csuklyás alak követett minket az erdőbe. |
| EN-AUG-1 | A hooded figure has regained much of his former strength. |
| HU-AUG-1 | Egy csuklyás alak szinte teljesen visszanyerte az erejét. |
| EN-AUG-2 | Sauron has followed us into the woods. |
| HU-AUG-2 | Szauron követett minket az erdőbe. |

**Table 3.3:** Augmentation via subtree swapping of subjects.

#### 3.1.2.3 Swapping with same predicate lemma

When constructing new sentences in either languages, two things are desired about the new, synthetic sentence: to be gramatically correct and to make some sense semantically. Grammatical correctness is hard to enforce, because it would require lots of hand-crafted, language-specific rules during the filtering for candidate sentences. We hypothesize however, that semantic correctness could be improved by only swapping sentences, which contain predicates that have the same lemma. The motivation behind this is that there are subjects and objects that appear around the same predicate more frequently and these are more likely to fit into the augmented sentence semantically, after performing subtree swapping. Having these constraints, example augmentations and the generated sentences are shown in Tables 3.4 and 3.5.

| | |
|---|---|
| Sentence1-EN | No one had seen my red bike since yesterday evening. |
| Sentence1-HU | Senki nem látta a piros biciklimet tegnap este óta. |
| Sentence2-EN | I see the fire in her eyes. |
| Sentence2-HU | Látom a tüzet a szemében. |
| EN-AUG-1 | No one had seen the fire since yesterday evening. |
| HU-AUG-1 | Senki nem látta a tüzet tegnap este óta. |
| EN-AUG-2 | I see my red bike in her eyes. |
| HU-AUG-2 | Látom a piros biciklimet a szemében. |

**Table 3.4:** Augmentation via subtree swapping of objects with same predicate lemma.

| | |
|---|---|
| Sentence1-EN | Nothing should be worth that. |
| Sentence1-HU | Semmi nem ér ennyit. |
| Sentence2-EN | Those two specimen are worth millions to the bio-weapons division. |
| Sentence2-HU | Az a két példány milliókat ér a biológiai fegyver részlegnek. |
| EN-AUG-1 | Nothing should be worth millions. |
| HU-AUG-1 | Semmi nem ér milliókat. |
| EN-AUG-2 | Those two specimen are worth that to the bio-weapons division. |
| HU-AUG-2 | Az a két példány ennyit ér a biológiai fegyver részlegnek. |

**Table 3.5:** Augmentation via subtree swapping of subjects with same predicate lemma.

#### 3.1.2.4 Predicate swapping

Given two sentence pairs of source and target sentences, we simultaneously swap the predicates of the sentences. An example augmentation and the generated sentences are shown in Table 3.6.

| | |
|---|---|
| Sentence1-EN | Everybody gets the rocket ship. |
| Sentence1-HU | Mindenki kap rakétát. |
| Sentence2-EN | Someone is hiding something. |
| Sentence2-HU | Valaki titkol valamit. |
| EN-AUG-1 | Everybody hiding the rocket ship. |
| HU-AUG-1 | Mindenki titkol rakétát. |
| EN-AUG-2 | Someone is gets something. |
| HU-AUG-2 | Valaki kap valamit. |

**Table 3.6:** Augmentation via predicate swapping.

## 3.2 Experiments

In this section I discuss the experiments carried out during my work. I trained transformer-based neural machine translation models for Hungarian-English and English-Hungarian with several types of data augmentation techniques.

### 3.2.1 Dataset

I used Hunglish2 as a training dataset for all my experiments, which is a sentence-aligned corpus containing more than 4 million Hungarian-English bisentences [52]. The raw corpus was scraped from the internet from a wide variety of sources, so it has coverage for several topics. The distribution of these subcorpora and their corresponding statistics are shown in Table 3.7.

| **Subcorpus** | **Tokens** | **Bisentences** |
|---|---|---|
| Modern literature | 37.1m | 1.67m |
| Classical literature | 17.2m | 652k |
| Movie subtitles | 3.2m | 343k |
| Software docs | 1.2m | 135k |
| Legal text | 56.6m | 1.351m |
| Total | 115.3m | 4.151m |

**Table 3.7:** Statistics of the Hunglish2 corpus

#### 3.2.1.1 Exploratory data analysis

Before the development of any model I performed exploratory data analysis on the Hunglish2 corpus. It is an essential step in carrying out any machine learning project, because on one hand it gives us insight into the dataset (and perhaps the problem that we are trying to solve) and on the other hand it provides useful information about the necessary preprocessing and data preparation steps that we will need to take. Thoughtful preprocessing is especially important in case of the Hunglish2 corpus, because parts of the data were crawled automatically from the internet, therefore the possibility of having erroneous sentences (e.g. HTML code) in the dataset is high. I performed all analyses in Jupyter Notebooks, using Pandas[1] for data manipulation and Seaborn[2] for data visualization.

[1] https://pandas.pydata.org/

[2] https://seaborn.pydata.org/

After downloading and loading the dataset to a Pandas dataframe, I first visualized the number of samples from each subcorpus (Figure 3.4).

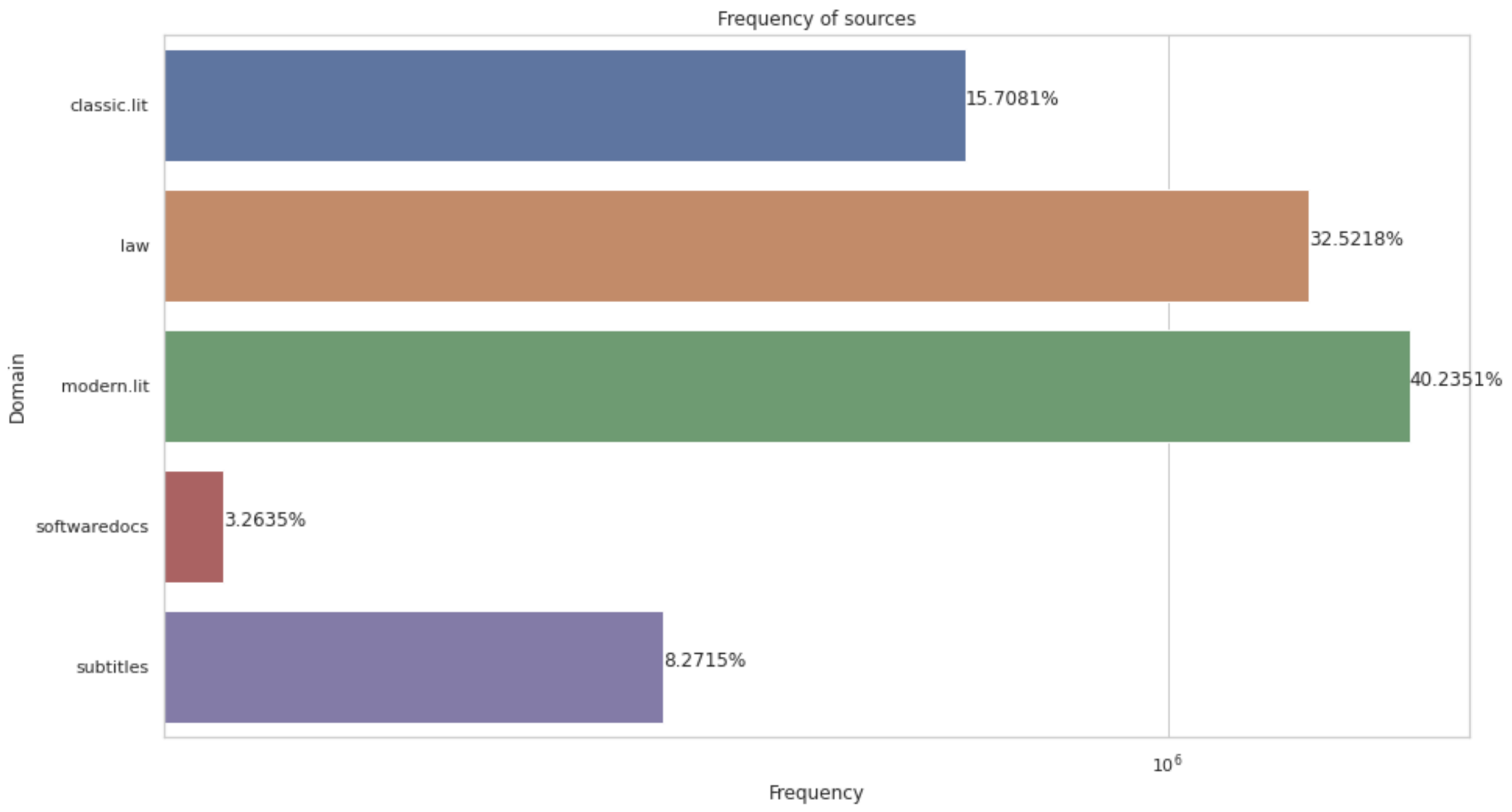

**Figure 3.4:** Distribution of subcorpora by domain in the Hunglish2 corpus.

I also examined length-based metrics of sentences individually for Hungarian and English. I tokenized all sentences using Spacy[3] and visualized word-level length distributions alongside the character-level distributions (Figure 3.5). Table 3.8 summarizes all length-based statistics, including max, min, mean, standard deviations and quantiles. Looking at these length-based metrics is an easy way to catch outliers, for example very long texts, that we likely want to remove.

| | | max | min | mean | q0.25 | q0.5 | q0.75 | q0.99 | q0.999 | stdev |
|---|---|---|---|---|---|---|---|---|---|---|
| Char | EN | 3997 | 0 | 63.57 | 25 | 46 | 84 | 269 | 470 | 57.81 |
| | HU | 3745 | 0 | 63.22 | 25 | 46 | 84 | 268 | 469 | 57.53 |
| Word | EN | 663 | 0 | 11.61 | 5 | 9 | 15 | 48 | 84 | 10.42 |
| | HU | 593 | 0 | 9.42 | 4 | 7 | 12 | 38 | 66 | 8.24 |

**Table 3.8:** Word- and character-level length metrics of the Hunglish2 corpus.

It is a common practice in NMT to compare the length-based metrics on the sentence pairs as well [42]. For example, if the ratio of number of tokens on the source and target sentences is very high, that could be an indicator of a false translation. Figure 3.6 shows the distribution of token count differences and token count ratios for both English and Hungarian. At this point, I removed some outlier sentence pairs from this plot ($x > q0.9999$), so the plots do not get distorted.

[3]https://spacy.io/

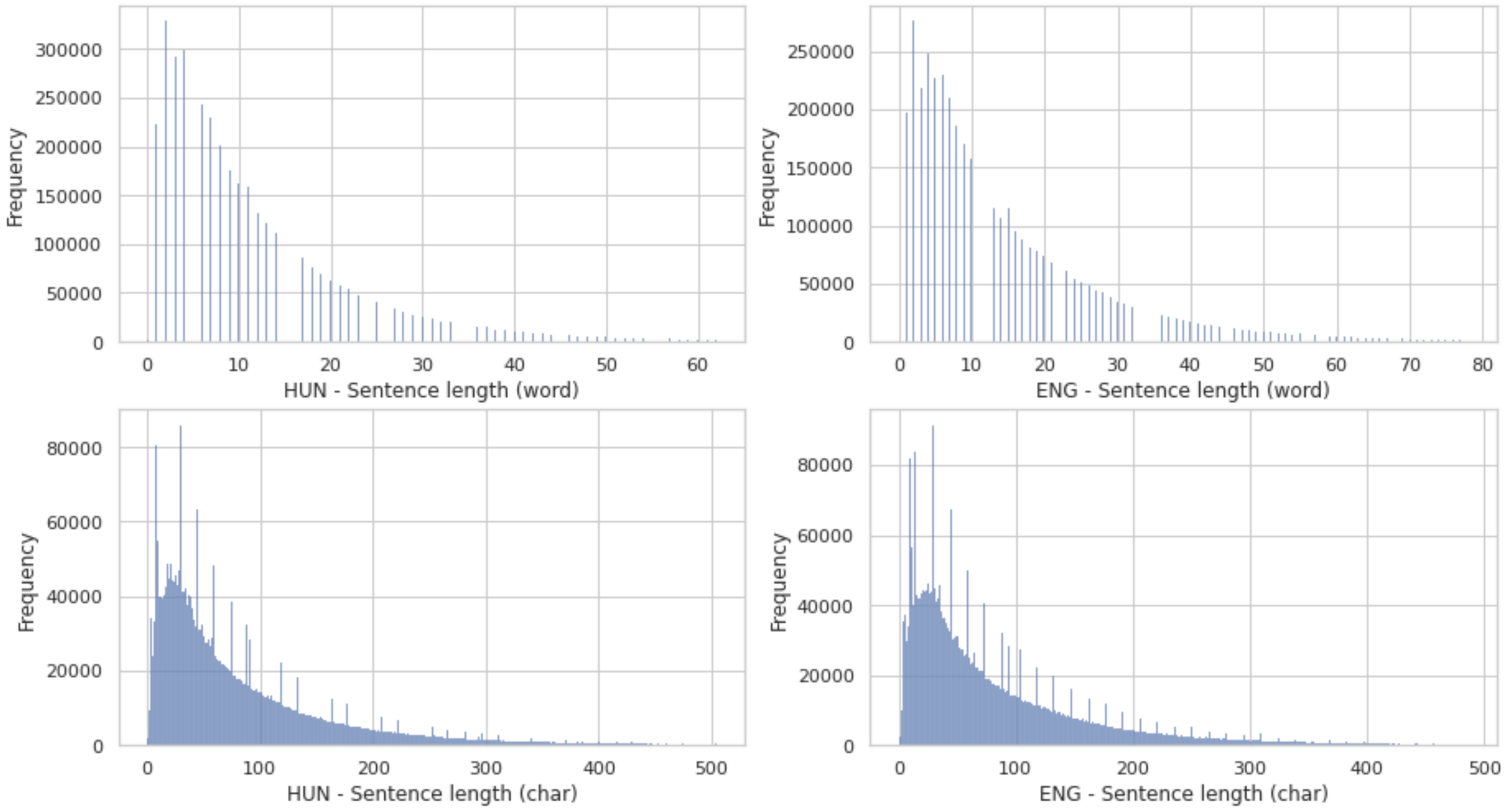

**Figure 3.5:** Word- and character-level length distributions of the Hunglish2 corpus.

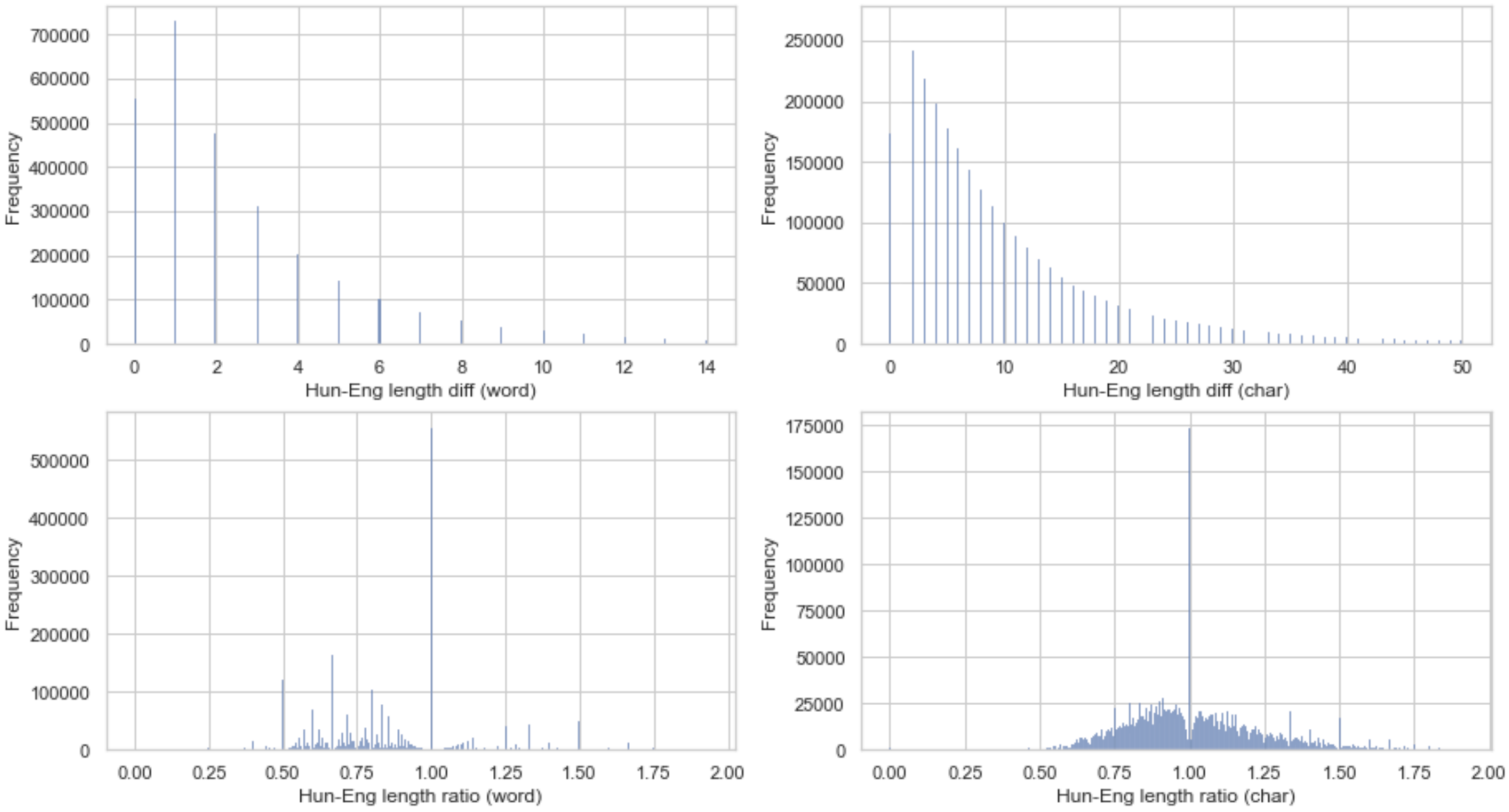

**Figure 3.6:** Word- and character-level distributions of the length difference and ratio between source and target sentences in the Hunglish2 corpus.

This information is useful during preprocessing, however applying these rules individually for filtering can be harmful, as there is a chance of removing data that should not be removed. If we set a threshold on the ratio of token counts between the source and target sentences to 1.8, short sentences could unnecessarily be removed. For example, the HU-EN sentence pair "Rendben" - "All right" would give a ratio of 2.0, although it is a perfect translation. For this reason, it is important to filter based on a combination of these metrics at the same time and the thresholds should be fine-tuned such that a significant amount of the original data remains after filtering. A good way to experiment with the thresholds is incrementally increasing/decreasing their value and looking at the portion of data remaining after removal. Plotting these charts can give us an idea about the optimal threshold values. The first experiment shown in Figure 3.7 shows the ratio of remaining sentences in the corpus if we remove sentence-pairs longer than an arbitrary word count.

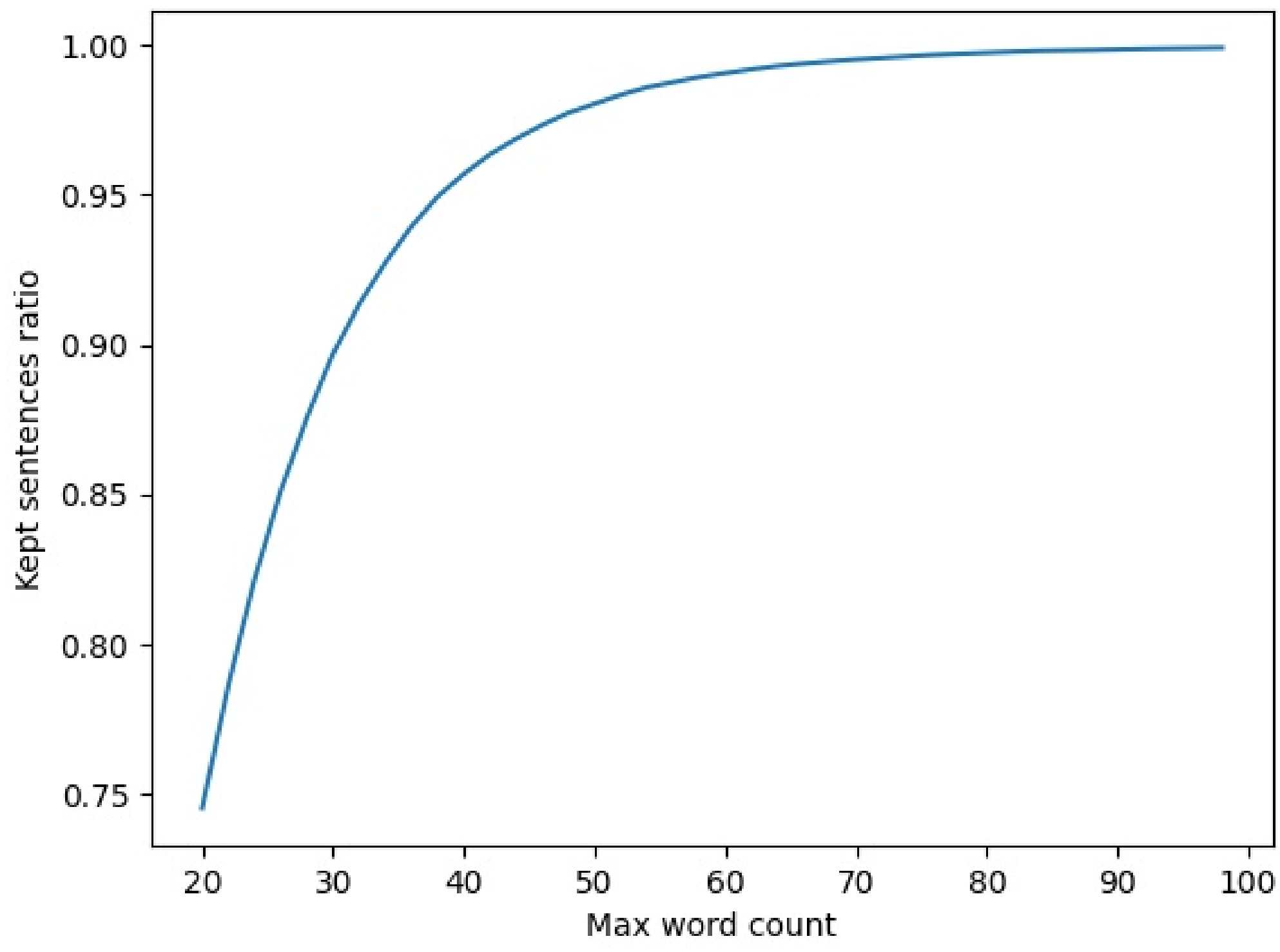

**Figure 3.7:** The effect of thresholding the maximum word count.

To avoid incorrectly removing samples as described in the example above, we experiment with having multiple filtering logics. We apply the ratio filtering in both cases, because it seems to be a very well working heuristic in prefiltering training data for neural machine translation. In the first experiment, along the ratio filtering, we apply different thresholds for the maximum word count as well (Figure 3.8 left). In the second experiment, we complement the ratio filtering with a threshold on the difference between the number of words in the source and target sentences (Figure 3.8 right).

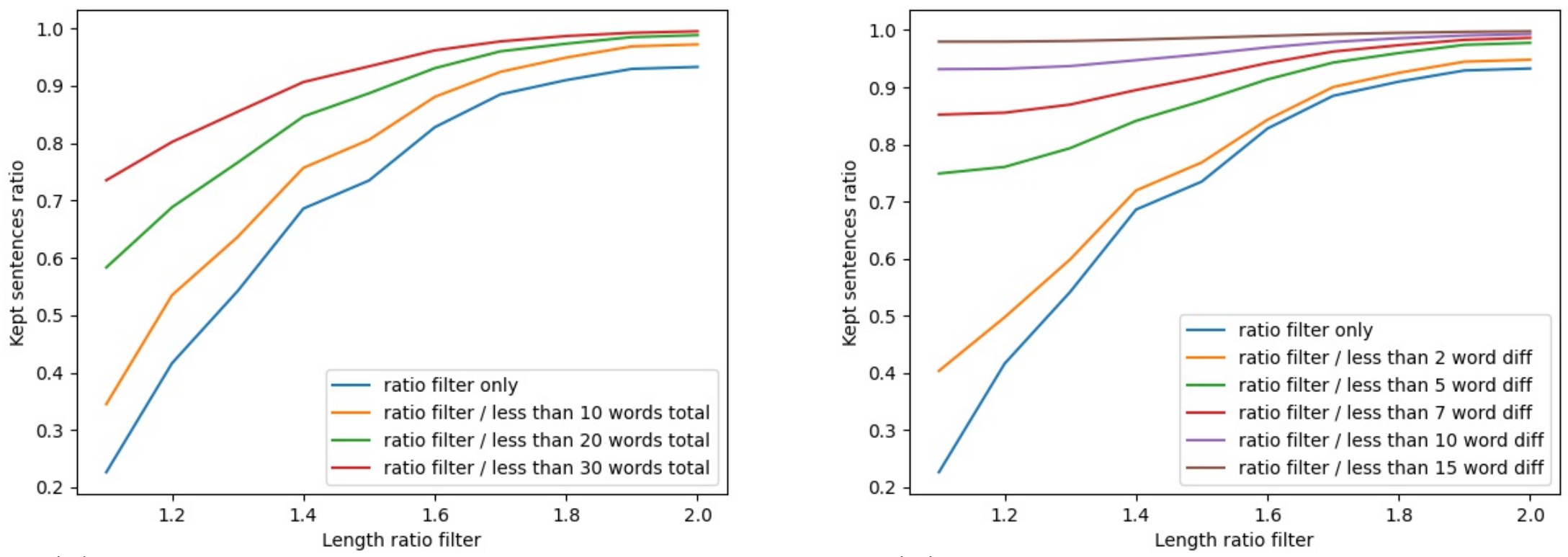

**(a)** Word ratio and total word count filtering **(b)** Word ratio and word difference filtering

**Figure 3.8:** Filtering experiments including word ratios and word counts. The left figure shows the effect of thresholding the word ratio with constant limits on a maximum word count. The right figure also shows the effect of thresholding the word ratio, but with constant limits on the word count difference between the source and target sentences.

#### 3.2.1.2 Preprocessing

Before performing any kind of length based filtering, it is important to perform preprocessing, because that might alter the length distributions. I first dropped all records where the source or target sentence was non existent or an empty string. A large number of sentences were wrapped in quotation marks without any specific reason. I did not remove these at first, but after training a few models, I noticed that the models really overfit to starting and ending sentences with quotation marks. For this reason, I removed all leading and trailing quotation marks. I also had to convert soft hyphens to regular hyphens, because the processing of soft hyphens caused a problem later during the generation of dependency parse trees. After this, based on the findings from the exploratory data analysis, I kept sentence-pairs that satisfied the following formula:

$$(0 < \text{word count} < 32) \wedge ((\text{word count difference} < 7) \vee (\text{word count ratio} < 1.6))$$

where the conditions are applied on both the source and target sentences respectively. After all the preprocessing and filtering steps, 3.4 million sentences remained in the dataset. The last thing that needs to be done before model development is splitting the dataset to train, development and test splits. As seen in Table 3.7, Hunglish2 is composed of subcorpora, which have significantly different linguistic characteristics, vocabulary and structure. Because of this, I applied stratified sampling, which means grouping the sentence pairs into subgroups called strata, based on the characteristics that they share: in this case the topic of the subcorpus. This could be sufficient for our cause, however luckily the files are structured in Hunglish2 such that a file contains sentence pairs from a single document. This is beneficial for us, because it allows stratification on a document level, meaning that the train, validation and test set will contain sentence pairs from all documents. The underlying motivaton is that some of the subcorpora still have huge diversity in its sentence pairs. For example, the classical literature subcorpus contains the works of William Shakespeare, Franz Kafka and the Bible as well, which significantly differ in a number of linguistic characteristics. To achieve this document-level stratified sampling, I created a simple Python script using scikit-learn[4]. The size of the final validation and test sets are 20k sentence pairs respectively, leaving all remaining samples as training data.

### 3.2.2 Implementation

In this section, I will discuss the implementation details of the augmentation methods and the model trainings. Figure 3.9 provides a high level overview of the entire pipeline developed for data augmentation and training.

#### 3.2.2.1 Augmentation

As the proposed augmentation methods introduced in this chapter rely on dependency parse trees, the first step is generating them for the entire training corpus both in Hungarian and English. Since millions of sentences are available, this takes several hours and therefore is worth precomputing in advance. I created a Python script that performs this precomputation and serializes the necessary information needed to construct the graphs into TSV files. These files include the words themselves, their lemmas and the respective dependency relations between them. To manipulate dependency trees, I created a generic framework in Python, which is essentially just a wrapper around the famous network analysis library Networkx[5]. I complemented the functionalities provided by Networkx with features, that could be useful for dependency parsing based data augmentation (e.g.

[4]https://scikit-learn.org/stable/
[5]https://networkx.org/

swapping subtrees, reconstructing sentences from the dependency tree etc.). The first step of generating augmented data is loading the precomputed dependency parse trees into these graphs. Next, we apply the filtering steps based on the graph representation of sentence-pairs (as discussed in Section 3.1.1). The filtering leaves about 63000 sentence pairs, which is roughly 2% of the total training dataset.

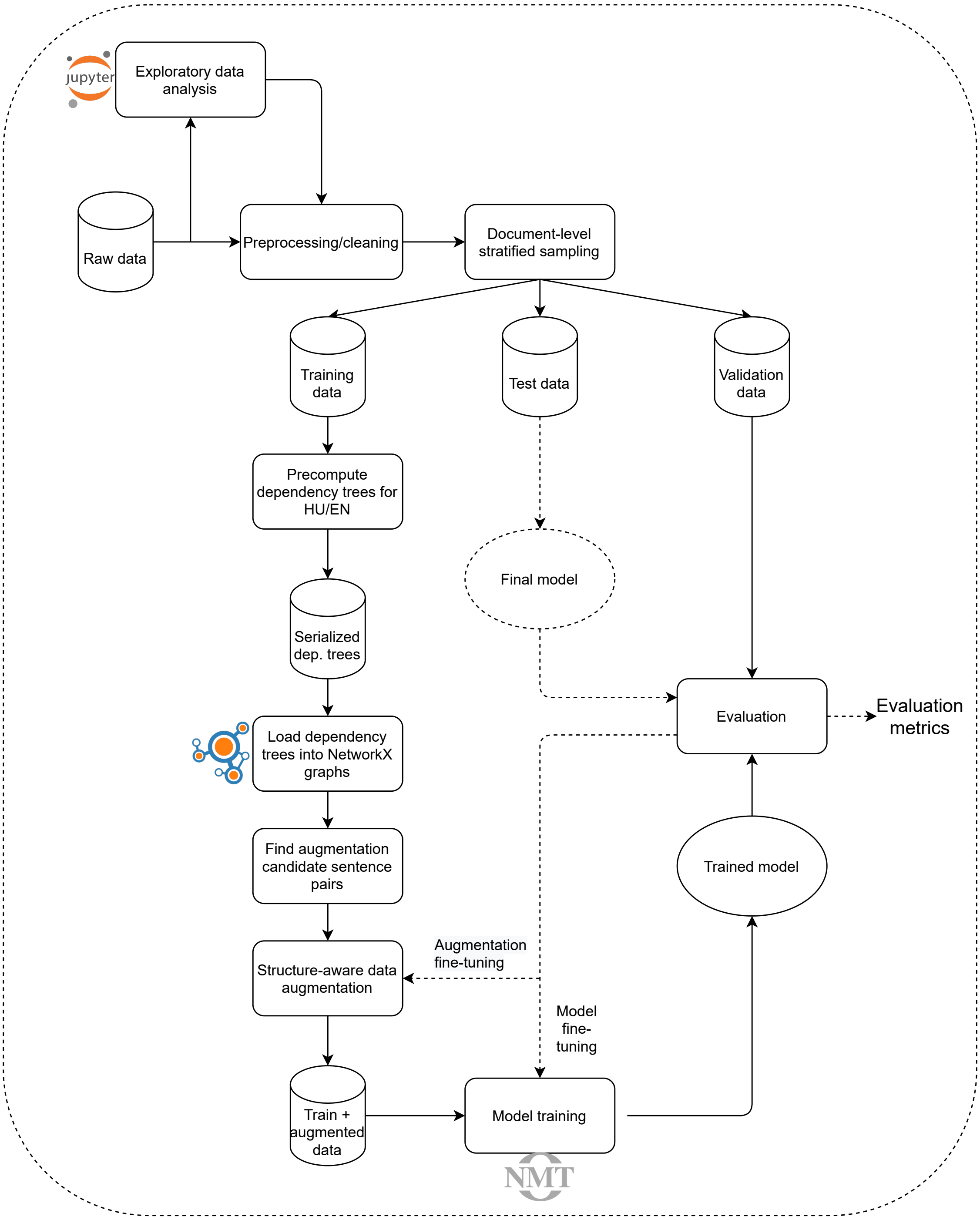

**Figure 3.9:** High-level overview of the complete pipeline of training neural machine translation models with augmented data.

Generating all sentences and then sampling a sufficient amount would be computationally infeasible, so given the amount of synthetic data that we want to generate, we sample a portion from the 63000 sentences that results in the desired number of new sentences after augmentation. The number of augmented sentences is usually given in ratio with respect to the size of the dataset. This ratio generally ranges from 0.1 to 1.0 depending on the augmentation method. For the experiments discussed in this thesis, I only evaluated augmentation methods with a 50% increment, due to time constraints. This value is a rule of thumb that generally worked in similar works, although it is important to acknowledge, that exploring other augmentation ratios could bring significant improvement to the proposed methods. For the simple object, subject and predicate swappings, this sampling is sufficient and all that is left is to perform the subtree swappings and reconstruct the sentences from the augmented graphs. When swapping subtrees among sentences that have the same predicate lemma, first we need to group all sentences based on the lemma of their predicates. After that we can sample pairs uniformly from each lemma group, to make sure that although we swap subtrees among sentences with the samme predicate lemma, we get diversity in the generated data. The most frequent lemma-pairs are visualized on Figure 3.10.

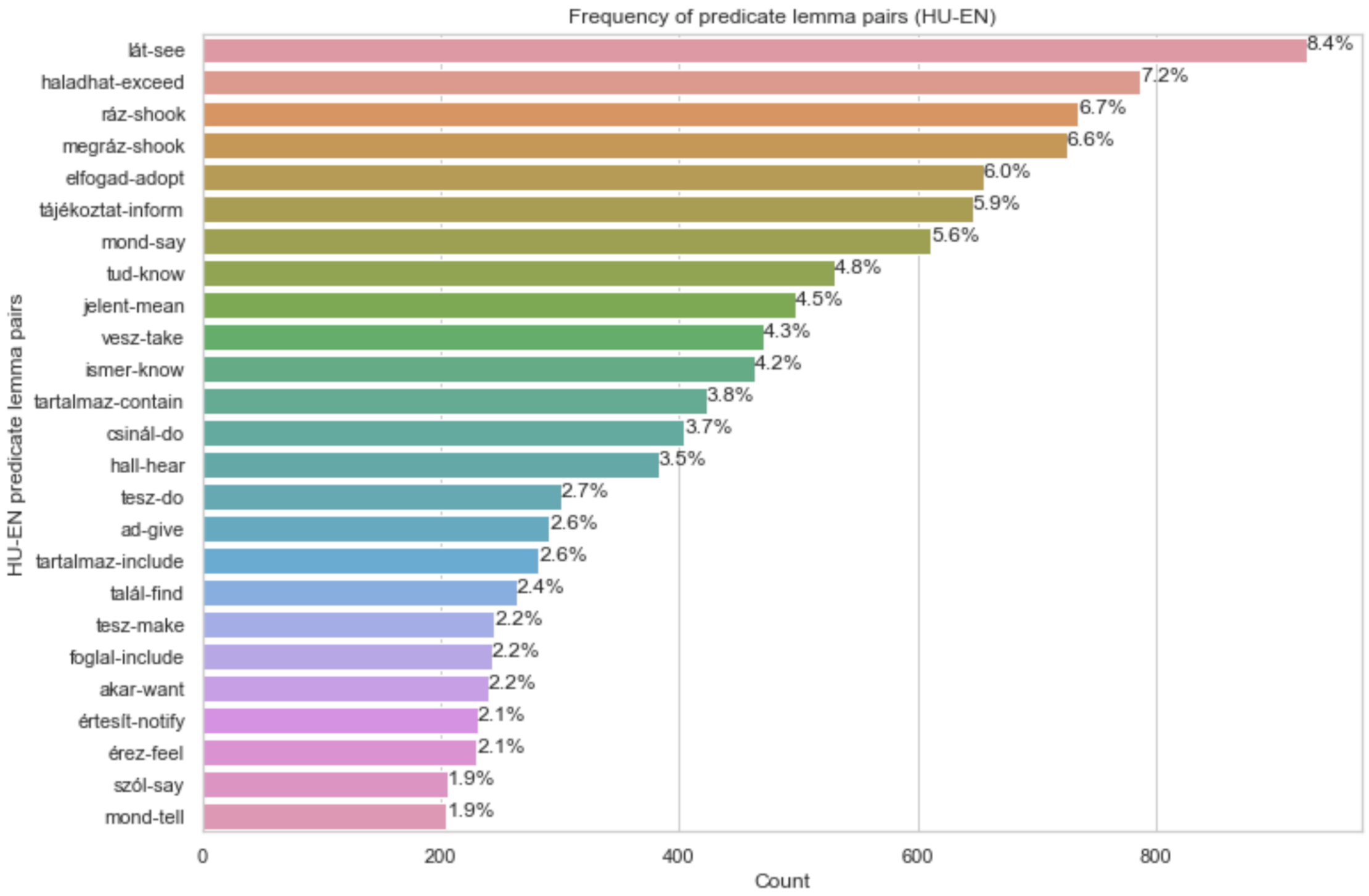

**Figure 3.10:** The 25 most frequent lemma-pairs in the Hunglish corpus.

When the sufficient amount of augmented data is ready for each method, all that is left is to shuffle them individually with the training set and start training models.

#### 3.2.2.2 Training

All models were trained using OpenNMT [25], which is an open-source neural machine translation framework. Machine translation generally requires a lot more data than most machine learning problems, therefore efficient data processing, data loading and training is inevitable. For this reason, OpenNMT prioritizes efficiency, while also being modular and extendable, supporting a wide variety of architectures.

For tokenizing the input sequences, I used the sentencepiece [29] subword tokenizer, which helps to keep the vocabulary lower in case of morphologically rich languages, like Hungarian. Each model that I trained was a transformer based encoder-decoder model, with identical parameters (Table 3.9). The trainings were performed on Nvidia V100 GPUs with each individual training lasting for about 18 hours.

| **Parameter** | **Value** |
|---|---|
| Word vector size | 512 |
| Batch size | 4096 tokens |
| Dropout | 0.1 |
| Optimizer | SparseAdam |
| Encoder type | transformer |
| Decoder type | transformer |
| Number of layers | 6 |
| Transformer FF dim | 2048 |
| LR warmup steps | 8000 |
| Learning rate | 2 |
| Decay method | Noam |
| Adam $\beta_2$ | 0.998 |

**Table 3.9:** Hyperparameters of the NMT models trained for evaluating the data augmentation.

# Chapter 4

# Results

## 4.1 Evaluation

The evaluation of models is performed using the Bilingual Evaluation Understudy (BLEU) score, which is a precision-oriented metric based on overlapping ngrams [40]. It was the first metric to show high correlation with human judgements of quality and therefore it has been the most common evaluation metric used in machine translation. Every metric shown in the tables in this section are reported for the validation dataset. The reason for this is to avoid data snooping, because this research project will carry on after the submission of my thesis and evaluating models on the test set now would induce bias in my decisions during future development. A transformer model without augmentations is considered a baseline for both languages. I compare this to the 5 proposed augmentation methods, along with the depth-based blanking method of Duan et al. [13], that I reimplemented as well. Each method is evaluated on models trained for 70k and 90k training steps respectively. These sum up to 14 models for Hungarian and for English, whose results are summarized in Tables 4.1 and 4.2. The purpose of this thesis was twofold: creating neural machine translation models both for Hungarian-English and English-Hungarian and developing data augmentation methods that presumably increase the performance of these models. For this reason I make the evaluation for the two objectives separately below.

| **Method** | **Augmentation ratio** | **Number of steps** | **BLEU** |
|---|---|---|---|
| Transformer baseline | - | 70k | 33.5 |
| | | 90k | 33.7 |
| Depth-based blanking [13] | 0.5 | 70k | **33.6** |
| | | 90k | 33.7 |
| Obj swapping | 0.5 | 70k | **33.8** |
| | | 90k | **33.9** |
| Subj swapping | 0.5 | 70k | 32.1 |
| | | 90k | 33.1 |
| Obj swapping same lemma | 0.5 | 70k | 32.7 |
| | | 90k | 32.9 |
| Subj swapping same lemma | 0.5 | 70k | 32.2 |
| | | 90k | 32.6 |
| Predicate swapping | 0.5 | 70k | 32.4 |
| | | 90k | **33.8** |

**Table 4.1:** Hungarian-English translation results

| **Method** | **Augmentation ratio** | **Number of steps** | **BLEU** |
|---|---|---|---|
| Transformer baseline | - | 70k | 27.6 |
| | | 90k | 28.6 |
| Depth-based blanking [13] | 0.5 | 70k | 27.5 |
| | | 90k | 27.9 |
| Obj swapping | 0.5 | 70k | 24.8 |
| | | 90k | 25.4 |
| Subj swapping | 0.5 | 70k | **27.8** |
| | | 90k | 27.3 |
| Obj swapping same lemma | 0.5 | 70k | 23.5 |
| | | 90k | 23.6 |
| Subj swapping same lemma | 0.5 | 70k | 27.3 |
| | | 90k | 27.4 |
| Predicate swapping | 0.5 | 70k | **27.8** |
| | | 90k | 27.9 |

**Table 4.2:** English-Hungarian translation results

### 4.1.1 Translation quality

The models in both experiments produce understandable translations, with or without augmentations. There has not been a substantial amount of work on neural machine translation for English-Hungarian. Tihanyi and Oravecz report a BLEU score of 26.13, although their model does not use the transformer architecture [51]. Our best model for English-Hungarian (Transformer baseline trained for 90k steps) achieves a BLEU score of 28.6. Table 4.3 shows three example translations by our best model.

| Example # | Sentence |
|---|---|
| 1 | **Source**<br>The lieutenant twitched.<br>**Reference**<br>A hadnagy csak forgolódott.<br>**Predicted**<br>A hadnagy összerezzent. |
| 2 | **Source**<br>And who is the magistrate who has reason to complain of the agent?<br>**Reference**<br>És ki az a köztisztviselő, akinek panasza lehet ön ellen?<br>**Predicted**<br>És ki az ügyész, akinek oka van panaszra? |
| 3 | **Source**<br>The Emperor was vexed, for he knew that the people were right; but he thought the procession must go on now!<br>**Reference**<br>A császár nagyon megütközött ezen; maga is úgy vélte, hogy igazat mondanak, de azt gondolta most már tovább kell mennem, nem futhatok haza szégyenszemre!<br>**Predicted**<br>A császár haragra gerjedt, mert tudta, hogy a népnek igaza van; de azt hitte, a menetnek mennie kell. |

**Table 4.3:** Example translations produced by the best English-Hungarian translator model.

The examples are grammatically correct in most cases and capture the meaning of the source sentence. It is worth noting, that in some cases the predicted translation feels to be of better quality than the reference (example #3 in Table 4.3). This is somewhat expected, because Hunglish2 is not a gold standard dataset, but an algorithmically aligned parallel corpus and there is some level of error in the alignment process.

For Hungarian-English we are not aware of any neural machine translation models published, therefore we compare every model to our transformer baseline. Our best model for Hungarian-English achieves a BLEU score of 33.9. It is a transformer complemented with object subtree swapping augmentation, trained for 90k steps. Table 4.4 shows three example translations by our best model.

| Example # | Sentence |
|---|---|
| 1 | **Source**<br>A hadnagy csak forgolódott.<br>**Reference**<br>The lieutenant twitched.<br>**Predicted**<br>The lieutenant kept turning. |
| 2 | **Source**<br>És ki az a köztisztviselő, akinek panasza lehet ön ellen?<br>**Reference**<br>And who is the magistrate who has reason to complain of the agent?<br>**Predicted**<br>And who is the civil servant who may complain of you? |
| 3 | **Source**<br>A császár nagyon megütközött ezen; maga is úgy vélte, hogy igazat mondanak, de azt gondolta most már tovább kell mennem, nem futhatok haza szégyenszemre!<br>**Reference**<br>The Emperor was vexed, for he knew that the people were right; but he thought the procession must go on now!<br>**Predicted**<br>The Emperor was shocked by this; he himself thought they were telling the truth, but he thought I must go on, not run home to shame! |

**Table 4.4:** Example translations produced by the best Hungarian-English translator model.

### 4.1.2 Augmentation quality

For Hungarian-English, object swapping and predicate swapping outperformed the baseline model with 33.9 and 33.8 BLEU scores respectively. It is worth noting however, that the other augmentation methods could contribute positively to the scores as well, perhaps with a different augmentation ratio. This is an important aspect to explore in future works. The reason why the augmentations do not work as expected could be either that the augmented sentences do not contain enough diversity (and are basically repetitive) or contain too many erroneous sentences. It is important to mention that dependency parsing is not perfect and a falsely generated dependency tree could result in wrong augmentations and therefore incorrect translations that divert model convergence. Furthermore, due to the different grammatical rules in the source and target languages, the subtrees that are swapped do not entirely correspond to the same entity (subject or object) within the sentence. This results in meaningless sentences. Table 4.5 shows two predictions on augmented samples by the best model (object swapping): #1 is a correctly augmented sentence, that is useful during training, while #2 is a less meaningful sample.

| Example # | Sentence |
|---|---|
| 1 | **Source**<br>A gondolat az eredetivel azonos sorszámot indukált benne.<br>**Reference**<br>The thought produced the same serial number.<br>**Predicted**<br>The thought induced the same serial number in it as the original. |
| 2 | **Source**<br>Vigyázz a kolostornak a temetkezéshez való jogát mondasz.<br>**Reference**<br>Watch the right of the monastery you say.<br>**Predicted**<br>Watch out for the monastery's right to be buried. |

**Table 4.5:** Example Hungarian-English translations on sentences that were generated by object swapping.

For English-Hungarian, subject swapping and predicate swapping outperformed the baseline model at 70k steps of training. At 90k steps however, the best model remained the transformer baseline. The same possible error causes are applicable for this language-pair as well, just like for Hungarian-English. Table 4.6 shows a correctly and a wrongly augmented sample using subject swapping.

The second example in Table 4.6 highlights two drawbacks of subject swapping. Firstly, it does not make much sense to swap subject subtrees in cases, where the coreference is not visible in the sentence pair. In the example, although we keep augmenting with the subject in Hungarian "szolgáltatók", the English equivalent will always contain "they". Some kind of coreference resolution could provide remedy for this, but with the current state of the algorithms, these samples should be excluded, because they distort training. Secondly, according to the subject-verb agreement, the verb in a sentence conjugates according to the subject. If we swap the subjects among sentences, the conjugations can get misaligned, which may result in false translations and grammatically incorrect sentences, such as in the example below: "A szolgáltatók tettem ezt?".

| Example # | Sentence |
|---|---|
| 1 | **Source**<br>The council gave him three brisk shakes.<br>**Reference**<br>A Tanács háromszor viharosan megrázta.<br>**Predicted**<br>A tanács három élénk rázkódást adott neki. |
| 2 | **Source**<br>did they do this?<br>**Reference**<br>A szolgáltatók tettem ezt?<br>**Predicted**<br>Ők tették ezt? |

**Table 4.6:** Example English-Hungarian translations on sentences that were generated by subject swapping.

## 4.2 Future work

Although some of the proposed methods outperformed the baseline model, there are several directions that are worth exploring in the future. First of all, more experiments need be performed with different parameters, specifically concerning the augmentation ratio. Secondly, there is room for more rigorous filtering of augmentation candidate sentences. I believe that the number of erroneous synthetic samples could be reduced by applying a more complex filtering logic, perhaps based on grammatical clues. Thirdly and lastly, I feel that this direction of data augmentation in NMT, that rely more on linguistics contain a lot of possibilities and it is worth experimenting with further manipulations over the dependency tree, other than what I presented in this work.

# Chapter 5

# Summary

I developed transformer-based neural machine translation models for Hungarian-English and English-Hungarian, complemented with structure-aware data augmentation methods that rely on dependency parsing. In the first part of my thesis I gave a thorough overview of neural networks, with a particular focus on architectures and algorithms that have had a huge impact on the field of natural language processing. As neural machine translation is a classical sequence to sequence problem, I put particular emphasis on the architectures that shaped the world of seq2seq, including recurrent neural networks, the attention mechanism and the transformer model. Data augmentation being one of the main objectives of this work, I highlighted the importance of augmentation in machine learning in general and reviewed the literature of augmentation in NLP. The contribution of my work is twofold. Firstly, I trained translations models for Hungarian-English and English-Hungarian that is capable of creating human-readable translations. Secondly, I proposed 5 structure-aware data augmentation methods, that are based on swapping specific subtrees simultaneously in the dependency trees of the source and target sentences. I close my thesis with ideas for potential enhancements in future works.

# Acknowledgements

I would like to thank Judit Ács for her endless support in guiding me through this work. I am grateful for both the scientific guidance and the hands-on supervision that she provided. I would also like to thank András Kornai for his useful comments and ideas on augmentation based on dependency parsing. Finally, I would like to thank my friends who also research NMT: Bence Bial, Balázs Frey and Patrick Nanys for their regular feedback on my ideas.